%% file: iclr2027_conference.tex
\documentclass{article} % For LaTeX2e

\usepackage{preprint,times}
\iclrpreprintcopy

\input{math_commands.tex}

\usepackage[utf8]{inputenc} % allow utf-8 input
\usepackage[T1]{fontenc}    % use 8-bit T1 fonts
\usepackage{hyperref}       % hyperlinks
\usepackage{url}            % simple URL typesetting
\usepackage{booktabs}       % professional-quality tables
\usepackage{amsfonts}       % blackboard math symbols
\usepackage{nicefrac}       % compact symbols for 1/2, etc.
\usepackage{microtype}      % microtypography
\usepackage{hyperref} 

\usepackage{booktabs}
\usepackage{adjustbox}
\usepackage{pdflscape}
\usepackage{rotating}

\usepackage{amsmath}
\usepackage{amssymb}
\usepackage{graphicx}
\usepackage{wrapfig}
\usepackage{sidecap}
\usepackage{multirow}
\usepackage[english]{babel}

\usepackage{float}
\usepackage{enumitem}
\usepackage{titlesec}
\usepackage{setspace}
\usepackage{enumitem}
\usepackage[table]{xcolor}
\usepackage{booktabs}
\setlist[itemize]{leftmargin=*}

\usepackage{amsthm}

\usepackage{algorithm}
\usepackage{algorithmic}

\newtheorem{proposition}{Proposition}

\definecolor{first}{HTML}{D9EAD3}   % light gold
\definecolor{second}{HTML}{D9EAF7}  % light green
\definecolor{third}{HTML}{FFF2CC}   % light blue

\hypersetup{hidelinks}

\author{
\textbf{Quanling Zhao} \quad
\textbf{Nilesh Prasad Pandey} \quad
\textbf{Ye Tian} \quad
\textbf{Tajana Rosing} \\
$^1$University of California San Diego \\
\texttt{\{quzhao,nppandey,yet002,tajana\}@ucsd.edu}
}

\begin{document}

\title{Superposed Inference for Hyperdimensional Computing}

\maketitle

\begin{abstract}
Hyperdimensional computing (HDC) is attractive for efficient and robust learning, but conventional inference still encodes every query independently, repeatedly paying the cost of high-dimensional projection. We introduce \textbf{SupHDC}, a new inference paradigm that processes multiple queries through a shared encoding computation. SupHDC assigns lightweight random slot keys, superposes the keyed queries before encoding, and uses slot-specific classifiers to recover their individual predictions. A random-feature kernel view explains why exact recovery of each hypervector is unnecessary: inference only needs to preserve the class evidence that determines the prediction. Across ten datasets, SupHDC achieves $1.39\times$ analytical speedup with no average accuracy loss, and up to $2.08\times$ speedup with only a $2.67$ percentage-point mean accuracy loss. On a Raspberry Pi~5, it delivers $2.01\times$ measured wall-clock speedup with a $2.26$ percentage-point loss in mean prediction accuracy. SupHDC shows that high-dimensional redundancy can be used not only for robustness, but also as capacity for shared inference.
\end{abstract}

\section{Introduction}
Hyperdimensional computing (HDC) has emerged as a promising paradigm for efficient and robust machine learning~\citep{kanerva2009hyperdimensional,kleyko2022survey}. By representing information with high-dimensional distributed vectors, HDC supports machine learning with simple arithmetic, lightweight training, associative inference, and offers strong tolerance to noise and computational imperfections~\citep{kanerva2009hyperdimensional,thomas2021theoretical}. These properties have motivated a rapidly growing body of work across different machine learning tasks and make HDC particularly attractive for TinyML, AI under harsh environments, and emerging hardware such as analog and in-memory computing~\citep{kleyko2023survey,kleyko2022vector,karunaratne2020memory}. In these settings, high dimensionality and distributed nature of HDC serve as a source of redundancy and robustness.
\begin{figure}[H]
    \centering
    \vspace{-3mm}
    \includegraphics[width=1\linewidth]{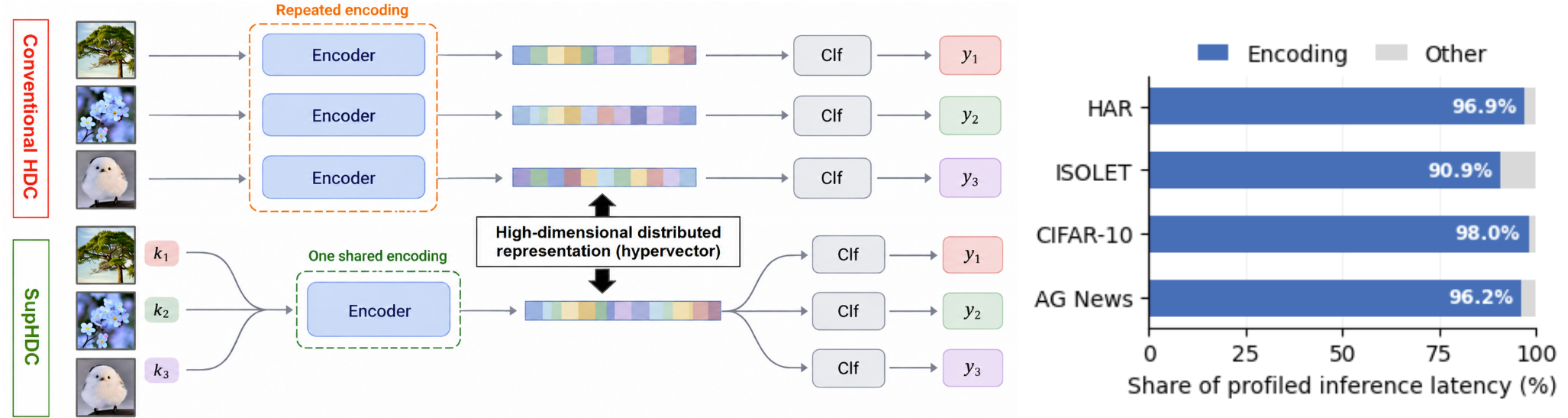}
    \vspace{-7mm}
    \caption{
    \textbf{Motivation for SupHDC.}
    (Left) Conventional HDC independently encodes each query into a high-dimensional distributed representation (hypervector), repeating the dominant encoding operation.
    SupHDC multiplexes multiple queries and amortizes the encoding cost through
    one shared encoding while retaining separate predictions.
    (Right) Profiling at $D=10{,}000$ shows that encoding dominates inference latency across different workloads, motivating SupHDC to amortize this cost.}
    \label{fig:intro_motivation}
    \vspace{-5mm}
\end{figure}

Despite HDC's attraction and many recent advances, conventional inference still processes each query independently. Every input must separately undergo the encoding operation. Processing $K$ queries therefore repeats essentially the same encoding pipeline $K$ times~\citep{verges2025classification,heddes2024hyperdimensional}. As illustrated in Fig.~\ref{fig:intro_motivation}(Left), this repeated encoding is the dominant inference cost in projection-based HDC systems. Our profiling in Fig.~\ref{fig:intro_motivation}(Right) shows that encoding accounts for the vast majority of inference latency across representative workloads. This observation is consistent with prior hardware studies that
identify encoding as a major HDC bottleneck~\citep{menon2023accelerating}. Reducing the representation dimension can lower this cost, but also removes the redundancy that makes HDC attractive in the first place~\citep{hernandez2024hyperdimensional,stock2024hyperdimensional}. This raises a different question: can high-dimensional redundancy be exploited not only for robustness, but also as capacity to process multiple inference queries through a shared encoding computation?

We introduce \textbf{SupHDC}, a framework that enables \textbf{superposed inference for hyperdimensional computing}. Given $K$ independent queries, SupHDC assigns each input a lightweight random slot key and superposes them before encoding. The mixed input is encoded only once, and slot-specific classifiers recover one prediction for each original query from the shared representation. Figure~\ref{fig:intro_motivation}(Left) highlights the key contrast: conventional HDC repeats encoding for every query, whereas SupHDC amortizes one encoding operation across multiple queries.

SupHDC is motivated by a simple observation: many projection-based HDC models can be viewed as approximations to kernel machines~\citep{thomas2021theoretical,zhao2025bridging}, so inference does not require exact recovery of each individual representation. It only requires the relevant class evidence to remain distinguishable. Based on this view, SupHDC assigns each query a lightweight random slot key, combines several keyed queries, and performs the expensive high-dimensional encoding only once for multiple samples. The resulting shared representation is interpreted by slot-specific classifiers, each recovering the prediction associated with its query. To compensate for residual interference between superposed queries, we adapt the slot classifiers directly to mixed inputs and selectively rerun only low-confidence predictions through the clean inference path. This design turns the superposition factor into a new efficiency knob: processing more queries together amortizes more encoding cost, while adaptation and selective fallback control the resulting accuracy loss.

Across ten datasets spanning sensing, vision, language, and industrial
applications, SupHDC demonstrates a favorable accuracy--compute tradeoff where it achieves $1.39\times$ analytical
inference speedup with no accuracy loss relative to conventional HDC model.
At a more aggressive operating point, the speedup increases to $2.08\times$
while incurring only a $2.67$ percentage-point mean accuracy loss. These
compute savings also translate to real hardware: on a Raspberry Pi~5,
SupHDC achieves $2.01\times$ measured wall-clock speedup with a $2.26$
percentage-point mean accuracy loss. Together with our theoretical analysis,
these results show that superposition can turn the redundancy of
high-dimensional representations into a practical source of shared inference
capacity, enabling a new inference paradigm for HDC and kernel
machines.

\section{Related Work}

\textbf{Hyperdimensional Computing(HDC):}
Hyperdimensional computing (HDC), also known as vector symbolic architectures (VSA), grew from cognitive and neuroscience-inspired models of distributed representation and computation\citep{plate1995holographic,kleyko2022survey}. Interest in HDC has expanded rapidly in recent years, with applications spanning biosignal processing, sensing, vision, language, genomics, graph learning, and other machine-learning and reasoning tasks\citep{kleyko2023survey,stock2024hyperdimensional,heddes2024hyperdimensional}. A theoretical view interprets many HDC-based learning model as randomized feature maps that approximate kernel machines\citep{zhao2025bridging,thomas2021theoretical}. This connection is particularly explicit for random Fourier feature encoders, where hypervector similarity approximates a shift-invariant kernel\citep{rahimi2007random}, and forms the basis of our analysis. A substantial body of prior work has also targeted the efficiency of HDC, including low-precision and quantized representations \citep{imani2019quanthd,pandey2025dpq}, as well as FPGA, in-memory, neuromorphic, and other specialized accelerators \citep{zhang2023hyperspikeasic,xu2024tri,angioli2024aeneashdc}. SupHDC addresses a different dimension of efficiency: rather than making each encoding cheaper, it reduces the number of independent encoding operations by amortizing one high-dimensional encoding across multiple queries. It is therefore complementary to both quantization and hardware acceleration.

\textbf{Superposition and Multiplexed Computation:}
Superposition has long been a fundamental operation in HDC and different vector symbolic architectures, where multiple representations are bundled into a fixed-dimensional representation that remains similar to its constituents \citep{plate1995holographic,kleyko2022survey}. In this classical use, superposition is primarily a mechanism for representing multiple pieces of information, often together with binding and permutation to encode structured objects. A separate line of work exploits unused representational capacity in neural networks themselves: model or parameter superposition stores multiple models within a shared parameter tensor and retrieves them using context-dependent transformations \citep{cheung2019superposition}. More closely related to our goal are methods that use superposition to reduce inference computation. DataMUX multiplexes several inputs through a neural network by transforming and combining them into one representation and learning a demultiplexing stage at the output \citep{murahari2022datamux}, while MIMONets use VSA-inspired binding and unbinding to process multiple inputs jointly through CNNs and Transformers, trading accuracy for throughput \citep{menet2023mimonets}. SupHDC takes a different route: it develops a kernel-machine view of superposed inference, showing how multiple HDC queries can share a single high-dimensional encoding itself while preserving information required for prediction. This reframes superposition from a mechanism for storing or demultiplexing representations into a principled means of amortizing the dominant cost.

\section{Preliminaries and Problem Definition}
\label{sec:prelim}

\textbf{Projection-Based HDC:}
For real-valued and tabular data, a widely used HDC encoder is a projection-based mapping parameterized by a random or quasi-random matrix $W\in\mathbb{R}^{D\times d}$, which maps an input $x\in\mathbb{R}^{d}$ into a high-dimensional representation $h(x)\in\mathbb{R}^{D}$ or $\mathbb{C}^{D}$~\citep{verges2025classification,hernandez2024hyperdimensional}. The randomized projection spreads information from each input dimension across many hypervector coordinates, producing the distributed representations that underlie HDC's robustness to noise and coordinate-level perturbations. Two common constructions are linear random-projection encoders, which apply a linear projection $Wx$ followed optionally by normalization or quantization, and nonlinear encoders based on random Fourier features (RFF), which apply a phase nonlinearity to the projected features. For the latter, we write $\phi_D(x)=D^{-1/2}[e^{i\omega_1^\top x},\ldots,e^{i\omega_D^\top x}]$, where the rows $\omega_j^\top$ of $W$ are drawn or quasi-randomly constructed from a prescribed spectral distribution~\citep{rahimi2007random}. In both cases, the projection from $d$ to a typically much larger dimension $D$ constitutes the principal encoding operation performed before lightweight HDC classification.

\textbf{HDC Classifiers as an Approximate Kernel Machine:}
For classification and associative-recall tasks, HDC typically stores encoded examples by bundling them into one or multiple prototype hypervectors. Generally speaking, for a class $c$, a prototype can be written as $P_c=\sum_{i:y_i=c}\alpha_i\phi_D(x_i)$, where $\phi_D$ denotes the HDC encoder and $\alpha_i$ is an optional sample weight. A query $x$ is then read out by comparing its encoded representation with the stored prototypes, for example through $s_c(x)=\langle \phi_D(x),P_c\rangle =\sum_{i:y_i=c}\alpha_i \langle\phi_D(x),\phi_D(x_i)\rangle$. Thus, whenever the encoder approximately preserves some similarity function $k(x,x')$, i.e., $\langle\phi_D(x),\phi_D(x')\rangle\approx k(x,x')$, the HDC readout becomes $s_c(x)\approx\sum_{i:y_i=c}\alpha_i k(x,x_i)$. The complete HDC classification pipeline can therefore be viewed as an explicit finite-dimensional approximation to a kernel machine. For linear random-projection encoders~\citep{zhao2025bridging}, the Johnson--Lindenstrauss property approximately preserves pairwise geometry and inner products, yielding a linear-kernel interpretation~\citep{achlioptas2003database,johnson1984extensions}. For nonlinear random-feature encoders, random Fourier features approximate a shift-invariant kernel through Bochner's theorem~\citep{rahimi2007random}. These two cases provide a common kernel-machine view of projection-based HDC, which we exploit later to analyze inference under superposition.

\textbf{Problem Definition:}
Consider $K$ independent inference queries $\{x_1,\ldots,x_K\}$. Conventional HDC classifier processes each query independently, requiring $K$ separate evaluations of the high-dimensional encoder $\phi_D(\cdot)$ before applying the lightweight prototype readout. When $D$ is large, this repeated encoding dominates inference cost. Our goal is therefore to replace these $K$ independent encoding operations with a shared computation while still producing one prediction for each original query.
Importantly, our goal is orthogonal to the choice of representation dimension $D$. For any selected $D$, SupHDC seeks to reduce the repeated cost of encoding multiple queries by sharing the projection computation. This is particularly useful when a larger $D$ is intentionally retained for representation fidelity or robustness, since SupHDC can reduce inference cost without requiring that representation to be compressed.
In this work, we focus on nonlinear RFF-based HDC, which provides a more expressive similarity model than linear random projection and can approximate a broad family of shift-invariant kernels.

\section{SupHDC: Superposed Inference for HDC}
\label{sec:method}

\begin{figure}
    \centering
    \includegraphics[width=1\linewidth]{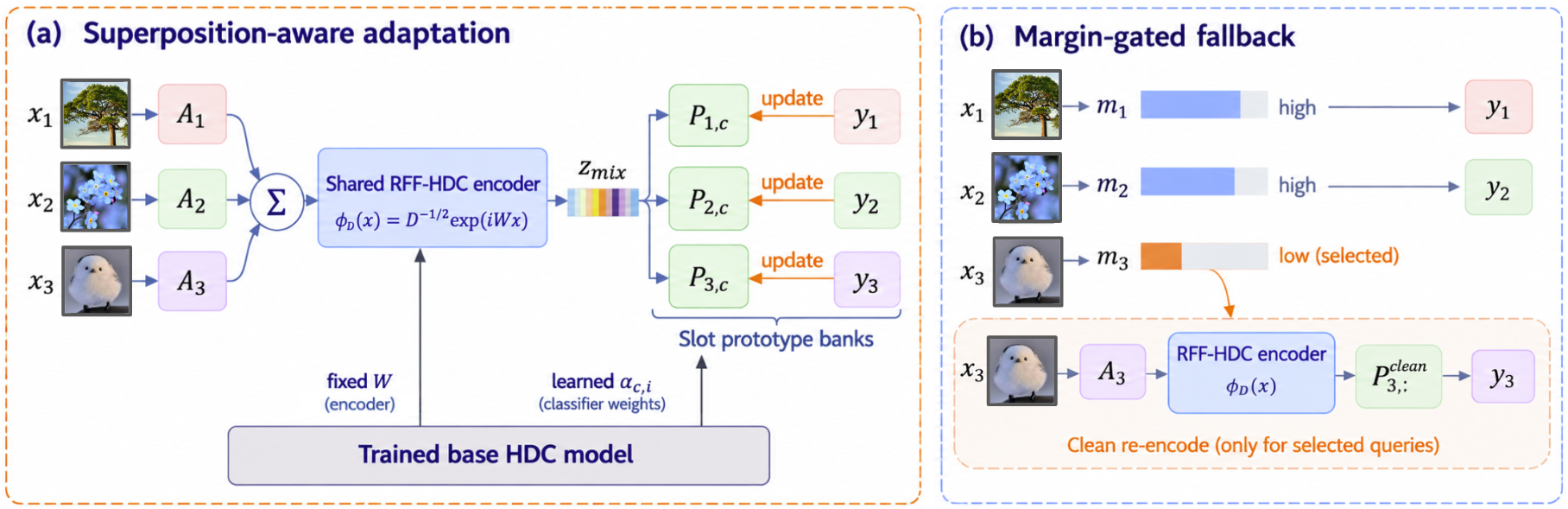}
    \vspace{-8mm}
    \caption{
    \textbf{SupHDC adaptation and selective fallback.}
    (a) Superposition-aware training adapts only the slot-specific prototype banks while keeping the base encoder fixed.
    (b) Margin-gated fallback selectively re-encodes low-confidence queries through the clean inference path.
    }
    \label{fig:method}
    \vspace{-7mm}
\end{figure}

We consider a trained nonlinear RFF-HDC classifier and target acceleration of
its inference stage. Let $W=[\omega_1,\ldots,\omega_D]^\top
\in\mathbb{R}^{D\times d}$ denote the fixed random-feature projection matrix.
For an input $x\in\mathbb{R}^{d}$, the encoder produces the complex
$D$-dimensional hypervector~\citep{yu2016orthogonal}
\begin{equation}
\small
    \phi_D(x)
    =
    \frac{1}{\sqrt{D}}
    \exp(iWx)
    =
    \frac{1}{\sqrt{D}}
    \left[
    e^{i\omega_1^\top x},\ldots,e^{i\omega_D^\top x}
    \right]
    \label{eq:rff_encoder}
\end{equation}
The base HDC model is trained conventionally, yielding a fixed encoder $W$
and class prototypes $\{P_c\}_{c=1}^{C}$ (or equivalently their learned
sample weights $\{\alpha_{c,i}\}$), with clean prediction
$\hat y=\arg\max_c \operatorname{Re}\langle\phi_D(x),P_c\rangle$.
SupHDC starts from this trained model. It leaves the high-dimensional encoder
unchanged and modifies how multiple inference queries are presented to and
read out from the shared encoder. Figure~\ref{fig:method} summarizes the two practical mechanisms used to reduce superposition interference: superposition-aware adaptation of the slot-specific readouts and margin-gated selective clean recomputation.

\textbf{Keyed Query Superposition:}
Consider $K$ independent queries $\{x_k\}_{k=1}^{K}$, where
$x_k\in\mathbb{R}^{d}$ and $d$ is the original input dimension.
SupHDC assigns each query to a fixed slot key
$A_k\in\mathbb{R}^{d\times d}$.
We define $A_k=S_k\Pi_k$, where $\Pi_k$ is a permutation matrix and
$S_k=\operatorname{diag}(s_{k,1},\ldots,s_{k,d})$ is a diagonal sign
matrix. We set $A_1=I_d$. For each remaining slot $k\geq2$,
$\Pi_k$ is sampled uniformly from the set of all $d$-dimensional
permutations, while the signs are sampled independently as
$s_{k,j}\sim\mathrm{Unif}\{-1,+1\}$ for $j=1,\ldots,d$.
The slot keys are sampled once and then kept fixed throughout training and
inference. Because both $\Pi_k$ and $S_k$ are orthogonal,
$A_k^\top A_k=I_d$. Keying therefore preserves the norm and geometry of each
query while placing different slots in randomized coordinate systems,
reducing cross-slot alignment. The keyed queries are then
superposed directly in the original input space:
\begin{equation}
\small
    x_{\mathrm{mix}}
    =
    \sum_{k=1}^{K} A_k x_k
    \label{eq:keyed_superposition}
\end{equation}
Unlike conventional HDC bundling, which combines representations only after
separate encodings have been computed, SupHDC performs superposition before
the expensive $d$-to-$D$ encoding. This allows the high-dimensional encoding
computation to be shared across all $K$ queries while adding only lightweight
operations in the original $d$-dimensional space.

\textbf{Shared Encoding and Slot-Specific Readout:}
After keying and superposition, SupHDC applies the high-dimensional encoder
only once. Importantly, for the nonlinear RFF encoder,
$\phi_D(\sum_k A_kx_k)\neq\sum_k\phi_D(A_kx_k)$ in general. The shared
representation is therefore not a simple summation of individually encoded
hypervectors that can be exactly recovered. SupHDC instead asks a weaker and
more useful question: can the shared representation retain enough
slot-specific class evidence to make the same predictions?
To extract this evidence, SupHDC associates each slot $k$ with its own
prototype bank $\{P_{k,c}\}_{c=1}^{C}$, where $C$ is the number of classes.
Let $\alpha_{c,i}$ denote the sample weight assigned by the underlying HDC
classifier to training example $x_i$ for class $c$. For each slot, we encode
the training examples under its key $A_k$ and construct
$\bar P_{k,c}=\sum_i\alpha_{c,i}\phi_D(A_kx_i)$, followed by
$P_{k,c}=\bar P_{k,c}/\|\bar P_{k,c}\|_2$.
Thus, each slot uses the same learned base-model sample weights, but constructs
its prototypes from examples encoded under the corresponding slot key. At inference, the
single shared representation is compared with all $K$ prototype banks:
\begin{equation}
\small
    z_{\mathrm{mix}}=\phi_D(x_{\mathrm{mix}}),\qquad
    \tilde{s}_{k,c}
    =
    \operatorname{Re}\langle z_{\mathrm{mix}},P_{k,c}\rangle,\qquad
    \tilde y_k
    =
    \arg\max_{c}\tilde{s}_{k,c}.
    \label{eq:shared_readout}
\end{equation}
Hence, one evaluation of the $D$-dimensional encoder produces the class scores
and prediction for every slot $k=1,\ldots,K$.

\textbf{Superposition-Aware Prototype Adaptation:}
The slot prototypes above are initialized from individually keyed examples,
whereas inference operates on the mixed representation
$z_{\mathrm{mix}}=\phi_D(\sum_{k=1}^{K}A_kx_k)$.
To account for the resulting superposition interference, we further adapt slot-specific prototype banks while keeping the encoder $\phi_D$ and the
keys $\{A_k\}_{k=1}^{K}$ fixed.
For each training step, we randomly sample a group
$\{(x_k,y_k)\}_{k=1}^{K}$, form
$x_{\mathrm{mix}}=\sum_{k=1}^{K}A_kx_k$, and compute a single shared
representation $z_{\mathrm{mix}}=\phi_D(x_{\mathrm{mix}})$.
Each slot $k$ then evaluates
$\tilde{s}_{k,c}=\operatorname{Re}\langle z_{\mathrm{mix}},P_{k,c}\rangle$
and predicts
$\hat y_k=\arg\max_c\tilde{s}_{k,c}$.
Thus, the same mixed representation provides $K$ supervised training signals,
one for each slot. Let $Q_{k,c}$ denote the working prototype, initialized as
$Q_{k,c}=P_{k,c}$. For a misclassified slot, $\hat y_k\neq y_k$, we update only the true and predicted class prototypes using
\begin{equation}
\small
Q_{k,y_k}
\leftarrow
Q_{k,y_k}
+\eta\bigl(1-\tilde{s}_{k,y_k}\bigr)z_{\mathrm{mix}},
\qquad
Q_{k,\hat y_k}
\leftarrow
Q_{k,\hat y_k}
-\eta\,\tilde{s}_{k,\hat y_k}z_{\mathrm{mix}}
\end{equation}
where $\eta$ is the adaptation learning rate.
The working prototypes are then normalized as
$P_{k,c}=Q_{k,c}/\|Q_{k,c}\|_2$.
This is the same score-dependent corrective update used by the underlying HDC
classifier, but applied directly to representations generated under
superposition.
Consequently, adaptation does not attempt to learn an inverse of the mixing
operation. It instead teaches each
slot-specific classifier to recognize its class evidence in the presence of
the interference it will encounter at inference time.
The original clean slot prototypes are kept unchanged and reserved for the
selective fallback mechanism which we described next.

\textbf{Margin-Gated Selective Fallback:}
Even after superposition-aware adaptation, some queries remain more vulnerable
to interference than others. SupHDC therefore uses the classification margin
to identify uncertain predictions and selectively spends additional computation
only on those queries. For slot $k$, let $\tilde{s}_{k,(1)}$ and
$\tilde{s}_{k,(2)}$ denote the largest and second-largest superposed class
scores, and define the confidence margin as
$m_k=\tilde{s}_{k,(1)}-\tilde{s}_{k,(2)}$.
A small margin means that only a small perturbation of the class scores could
change the predicted label, making that query a natural candidate for clean
recomputation. Because these scores are already produced by the slot-specific
readout, the confidence test itself adds negligible overhead. For each inference batch, suppose there are $N_b$ valid slot predictions.
Given a fallback ratio $q\in(0,1)$, SupHDC selects the
$\lceil qN_b\rceil$ predictions with the smallest margins in that batch.
For each selected query in slot $k$, we recompute its clean keyed encoding,
$z_k^{\mathrm{clean}}=\phi_D(A_kx_k)$,
then classify it using the corresponding clean slot prototype bank,
$s^{\mathrm{clean}}_{k,c}
=\operatorname{Re}\langle z_k^{\mathrm{clean}},
P^{\mathrm{clean}}_{k,c}\rangle$, and replace its superposed prediction.
All unselected queries retain their original superposed predictions.
Thus, fallback does not attempt to estimate or cancel interference. It simply
re-encodes the low-confidence queries for which interference is most likely to
change the decision. The fallback ratio $q$ therefore provides a direct
accuracy--throughput knob at inference time.

\textbf{Computational and Storage Cost:}
SupHDC reduces inference cost by amortizing the expensive high-dimensional
encoding across multiple queries. Let $E$ denote the cost of one encoding,
$R$ the cost of one prototype readout, and $F_{\mathrm{key}}$ the lightweight
cost of signed permutations and summation. Processing $K$ queries independently
costs $K(E+R)$, whereas SupHDC costs approximately
$E+KR+F_{\mathrm{key}}$. Thus, SupHDC removes repeated encoding while
retaining the $K$ slot-specific readouts, so its benefit is largest when
encoding dominates the inference pipeline, which is exactly the bottleneck observed in many HDC systems. With a fallback ratio $q$, approximately $qK$ queries per group incur an
additional clean encoding and readout, giving a cost of roughly
$E+KR+F_{\mathrm{key}}+qK(E+R)$. The fallback ratio therefore trades part of
the compute saving for improved prediction fidelity. Storage remains modest
because the encoder matrix is shared across all slots. The additional storage
comes mainly from the slot-specific prototype banks and lightweight keys.

\section{Understanding Superposed Inference}
\label{sec:theory}

We analyze the core superposed inference before prototype adaptation; the
adaptation and fallback mechanisms in section~\ref{sec:method} are additional
mechanisms for handling the interference characterized below.

\textbf{Kernel View of a Superposed Slot:}
To understand why SupHDC can recover useful predictions from a mixed input,
consider a particular slot $k$. Let $u_k=A_kx_k$ denote the keyed query of
interest and let
$v_k=\sum_{\ell\neq k}A_\ell x_\ell$ denote the aggregate contribution from
the remaining $K-1$ slots, so that
\begin{equation}
\small
    x_{\mathrm{mix}}
    =
    u_k+v_k,
    \qquad
    u_k=A_kx_k,
    \qquad
    v_k=\sum_{\ell\neq k}A_\ell x_\ell
    \label{eq:slot_decomposition}
\end{equation}
From the kernel-machine view in section~\ref{sec:prelim}, the relevant question
is not whether the individual representation $\phi_D(u_k)$ can be exactly
reconstructed from $\phi_D(u_k+v_k)$. Instead, we ask how the additional
term $v_k$ perturbs the kernel similarities, and hence the class scores, used
to classify slot $k$. For a shift-invariant kernel
$k(x,x')=\kappa(x-x')$, this reduces to understanding how replacing a clean
difference $\delta$ by $\delta+v_k$ changes the kernel value
$\kappa(\delta)$.

\textbf{Score Preservation under Superposition:}
Ignoring finite-$D$ random-feature approximation for the moment, consider slot
$k$ and one of its keyed training examples $A_kx_i$.
Without superposition, their kernel similarity depends on
$\delta_{k,i}=A_k(x_k-x_i)$ and is given by
$\kappa(\delta_{k,i})$.
After adding the other slots, the query becomes $u_k+v_k$, so the same
similarity becomes $\kappa(\delta_{k,i}+v_k)$.
We separate this perturbed similarity into a common scaling term and a
residual,
$\kappa(\delta+v)=\kappa(v)\kappa(\delta)+r_\kappa(\delta,v)$.
Writing the normalized prototype score as a weighted kernel sum, with
prototype normalization absorbed into coefficients $\beta_{k,c,i}$, summing
over the training examples that form a class prototype gives
\begin{equation}
\small
    \tilde{s}_{k,c}
    =
    \rho_k s_{k,c}
    +
    \epsilon_{k,c},
    \qquad
    \rho_k=\kappa(v_k),
    \qquad
    \epsilon_{k,c}
    =
    \sum_i \beta_{k,c,i}
    r_\kappa(\delta_{k,i},v_k)
    \qquad \textit{(See Appendix~\ref{app:score_decomposition} for proof)}
    \label{eq:score_decomposition}
\end{equation}
Here, $s_{k,c}$ is the clean class score and $\epsilon_{k,c}$ collects the
remaining class-dependent distortion caused by superposition.
This decomposition highlights the key point: $\rho_k$ affects all classes in
slot $k$ in the same way, whereas $\epsilon_{k,c}$ can change their relative
ordering. Therefore, when $\rho_k>0$, accurate superposed inference does not
require the mixed scores to match the clean scores exactly; it only requires
the class-dependent distortion to remain small enough that the winning class
is preserved.

\textbf{Random Keys Suppress the Residual:}
For a normalized real-valued shift-invariant positive-definite kernel,
Bochner's theorem gives
$\kappa(t)=\mathbb{E}_{\omega}[e^{i\omega^\top t}]$. The residual
$r_\kappa(\delta,v)
=\kappa(\delta+v)-\kappa(\delta)\kappa(v)$
therefore measures the centered spectral interaction between the desired
displacement $\delta$ and the interference $v$. Random slot keys randomize
the relative coordinate alignment of these two terms. For the RBF kernel,
this relationship becomes explicit:
\begin{equation}
\small
    r_\kappa(\delta,v)
    =
    \kappa(\delta)\kappa(v)
    \left[
    \exp\!\left(-\frac{\delta^\top v}{\sigma^2}\right)-1
    \right]
    \qquad \textit{(See Appendix~\ref{app:rbf_residual} for proof)}
    \label{eq:rbf_residual}
\end{equation}
Hence, the residual is directly controlled by the cross-slot alignment
$\delta^\top v$. In SupHDC,
$\delta=A_k(x_k-x_i)$ and
$v_k=\sum_{\ell\neq k}A_\ell x_\ell$, so this cross-term is a sum of
interactions involving the random relative keys $A_k^\top A_\ell$.

\begin{proposition}[Random-key cross-slot alignment]
\label{prop:key_alignment}
Fix slot $k$ and condition on the input vectors. Let
$a=x_k-x_i$, $\delta=A_ka$, and
$v_k=\sum_{\ell\neq k}A_\ell x_\ell$, where
$A_1=I$ and $A_2,\ldots,A_K$ are independently sampled uniform
signed-permutation matrices. Then, with expectation taken over the random
keys,
\begin{equation}
\small
    \mathbb{E}\!\left[\delta^\top v_k\right]=0,
    \qquad
    \mathbb{E}\!\left[(\delta^\top v_k)^2\right]
    =
    \frac{\|a\|_2^2}{d}
    \sum_{\ell\neq k}\|x_\ell\|_2^2
    \qquad \textit{(See Appendix~\ref{app:proof_key_alignment} for proof)}
    \label{eq:key_alignment_moment}
\end{equation}
Consequently, for any $\eta\in(0,1)$, with probability at least
$1-\eta$,
\begin{equation}
\small
    |\delta^\top v_k|
    \leq
    \frac{\|a\|_2}{\sqrt{d}}
    \sqrt{
        \frac{\sum_{\ell\neq k}\|x_\ell\|_2^2}{\eta}
    }
    \qquad \textit{(See Appendix~\ref{app:proof_key_alignment} for proof)}
    \label{eq:key_alignment_prob}
\end{equation}
\end{proposition}

Proposition~\ref{prop:key_alignment} makes this effect explicit. If the
interfering inputs have comparable norm $\|x_\ell\|_2\approx r$, then
$ \sqrt{\mathbb{E}\!\left[(\delta^\top v_k)^2\right]} \approx \|a\|_2 r \sqrt{\frac{K-1}{d}}$. Thus, relative to the scale $\|a\|_2r$, the typical cross-slot interaction
grows as $\sqrt{K-1}$ but decreases as $1/\sqrt{d}$. Increasing $K$
therefore increases the interference load, while, for fixed vector norms,
increasing $d$ reduces coherent cross-slot alignment. For the RBF residual in equation~\ref{eq:rbf_residual}, using
$|e^{-t}-1|\leq e^{|t|}|t|$ gives
\begin{equation}
\small
    |r_\kappa(\delta,v)|
    \leq
    \kappa(\delta)\kappa(v)
    \exp\!\left(
        \frac{|\delta^\top v|}{\sigma^2}
    \right)
    \frac{|\delta^\top v|}{\sigma^2}
    \qquad \textit{(See Appendix~\ref{app:rbf_residual} for proof)}
    \label{eq:rbf_residual_bound}
\end{equation}
Thus, when the cross-slot alignment is small relative to the kernel scale
$\sigma^2$, the residual is correspondingly small. Since
$|\epsilon_{k,c}|
\leq
\sum_i |\beta_{k,c,i}|
|r_\kappa(\delta_{k,i},v_k)|$,
Proposition~\ref{prop:key_alignment} connects the geometry induced by random
keying to class-dependent score distortion, with the resulting sensitivity
also determined by the kernel bandwidth.

\textbf{Margin-Based Prediction Preservation:}
The score decomposition gives a direct sufficient condition for preserving
the clean prediction.

\begin{proposition}[Prediction preservation under superposition]
\label{prop:prediction_preservation}
Let
$\tilde{s}_{k,c}=\rho_k s_{k,c}+\epsilon_{k,c}$ with $\rho_k>0$.
Assume the clean classifier has a unique prediction
$y_k=\arg\max_c s_{k,c}$, with margin
$\gamma_k=s_{k,y_k}-\max_{c\neq y_k}s_{k,c}>0$.
If
$2\|\epsilon_k\|_\infty<\rho_k\gamma_k$,
then
$\arg\max_c \tilde{s}_{k,c}=y_k$. $\textit{(See Appendix~\ref{app:proof_prediction_preservation} for proof)}$

\end{proposition}

Proposition~\ref{prop:prediction_preservation} shows that a positive common
scaling does not change class ordering by itself; what matters is whether the
class-dependent distortion is small relative to the scaled clean margin
$\rho_k\gamma_k$. Consequently, large-margin queries can tolerate more
superposition-induced distortion, whereas small-margin queries are easier to
flip. This observation also motivates the margin-gated fallback in
section~\ref{sec:method}, which uses the observed top-two superposed score margin
as a practical indicator of prediction vulnerability.

\textbf{Finite-$D$ Random-Feature Approximation:}
The analysis above describes the underlying kernel scores, whereas SupHDC
implements them using a finite $D$-dimensional random-feature map.
For standard i.i.d. random Fourier features, the empirical kernel
$\hat{k}_D(x,x')=\operatorname{Re}\langle\phi_D(x),\phi_D(x')\rangle$
converges to the target kernel as $D$ increases, with pointwise approximation
error of order $O(D^{-1/2})$~\citep{sutherland2015error}. The superposed score can therefore be viewed as:
\begin{equation}
\small
    \tilde{s}^{(D)}_{k,c}
    =
    \rho_k s_{k,c}
    +
    \epsilon^{\mathrm{sup}}_{k,c}
    +
    \epsilon^{\mathrm{RFF}}_{k,c}
    \qquad \textit{(See Appendix~\ref{app:finiteD} for proof)}
    \label{eq:finiteD}
\end{equation}
where $\epsilon^{\mathrm{sup}}_{k,c}$ denotes the structural interference
introduced by superposition and $\epsilon^{\mathrm{RFF}}_{k,c}$ denotes the
finite-feature approximation error. These terms have different origins:
$K$ controls how many interfering queries are superposed, the input dimension
$d$ controls random-key separation for fixed input scale, and $D$ controls
how accurately the finite representation approximates the underlying kernel.

\section{Experiments}
\label{sec:experiments}

\begin{figure}[t]
    \centering
    \includegraphics[width=1\linewidth]{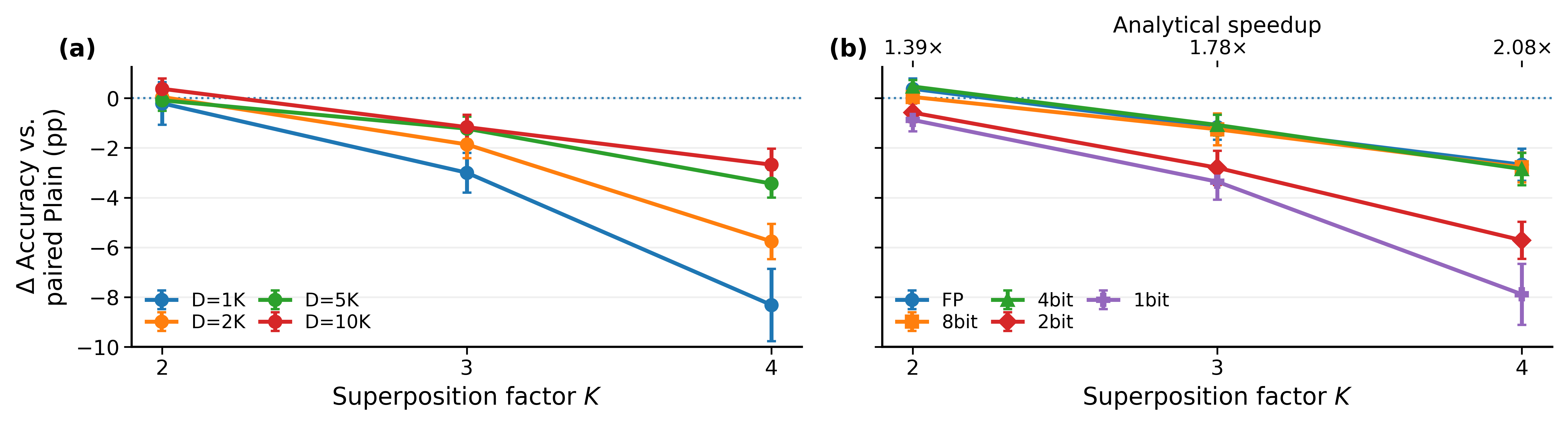}
    \vspace{-9mm}
    \caption{
    \textbf{SupHDC provides an efficiency knob complementary to representation
    dimension and numerical precision.}
    (a) Accuracy change of SupHDC with prototype adaptation and 20\%
    margin-gated fallback across representation dimensions.
    Larger $D$ supports more aggressive superposition with smaller accuracy
    degradation.
    (b) SupHDC under different numerical precisions, always compared against
    Plain HDC at the same precision. The upper axis reports the analytical
    speedup introduced by SupHDC, showing that quantization and superposed
    inference can be applied jointly.
    }
    \label{fig:dimension_quantization}
    \vspace{-4mm}
\end{figure}

\textbf{Experimental Setup:}
We evaluate SupHDC on 10 datasets: HAR~\citep{anguita2013public}, ISOLET~\citep{cole1990isolet}, CIFAR-10~\citep{krizhevsky2009learning}, Airline Sentiment~\citep{rane2018sentiment}, AG News~\citep{zhang2015character}, IMDB~\citep{maas2011learning}, 20 Newsgroups~\citep{lang1995newsweeder}, SECOM~\citep{secom_179}, MNIST~\citep{lecun1998gradient}, and Fashion-MNIST~\citep{xiao2017fashion}, spanning sensing, vision, language, and industrial applications. The main evaluation
covers hypervector dimensions $D\in\{1{,}000,2{,}000,5{,}000,10{,}000\}$, superposition factors
$K=2$--$10$, and numerical quantization levels $\{\mathrm{FP},8,4,2,1\}$ bits, with
results averaged over five random seeds. We compare conventional RFF-HDC
(Plain) against SupHDC, superposition-aware adaptation, margin-gated fallback,
and their combination. All inputs are $\ell_2$-normalized after
dataset-specific preprocessing, controlling input scale as in the analysis of
section~\ref{sec:theory}. Accuracy is our primary predictive metric. 
Computational efficiency is measured using a hardware-independent cost
model that counts projection FLOPs and prototype-scoring
FLOPs. Analytical speedup is the ratio of the paired Plain HDC cost to the
corresponding SupHDC cost. For $K$ queries, Plain requires $K$ projections
whereas SupHDC requires one shared projection; fallback additionally includes
clean recomputation of the selected queries.
We additionally deploy SupHDC on
a Raspberry Pi~5 under single-thread CPU execution as a resource-constrained
edge evaluation representative of TinyML-oriented deployment.

\textbf{Accuracy--Compute Tradeoff:}
We first ask whether SupHDC can reduce inference computation while preserving
the predictive behavior of Plain HDC. Table~\ref{tab:main_grid} summarizes the
main results at $D=10{,}000$ and full precision. Prototype adaptation alone provides aggressive compute reduction, reaching $1.92\times$, $2.78\times$, and $3.58\times$ analytical speedup for $K=2,3,4$, respectively, but incurs
\begin{wraptable}{r}{0.55\columnwidth}
    \vspace{-1.7em}
    \centering
    \scriptsize
    \setlength{\tabcolsep}{2.8pt}
    \caption{
    \textbf{Accuracy--compute tradeoff} at $D=10{,}000$ and full precision.
    Adapt. denotes prototype adaptation and FB denotes 20\% margin-gated
    fallback. $\Delta$ Acc. is relative to Plain. $\pm$ denotes standard error of the mean(SEM) across dataset-level means.
    }
    \label{tab:main_grid}
    \vspace{1em}
    \begin{tabular}{lcccc}
        \toprule
        Method & $K$ & Acc. (\%) & $\Delta$ Acc. (pp) & Analyt. speedup \\
        \midrule
        Plain
        & 1 & 84.39 & 0.00 & $1.00\times$ \\
        \midrule

        SupHDC + Adapt.
        & 2 & 83.30 & $-1.09\pm0.81$ & $1.92\times$ \\

        \textbf{+ FB}
        & 2 & \textbf{84.76} & $\mathbf{+0.37\pm0.42}$ & $1.39\times$ \\

        SupHDC + Adapt.
        & 3 & 80.17 & $-4.21\pm0.87$ & $2.78\times$ \\

        \textbf{+ FB}
        & 3 & \textbf{83.22} & $\mathbf{-1.16\pm0.51}$ & $1.78\times$ \\

        SupHDC + Adapt.
        & 4 & 77.27 & $-7.11\pm0.97$ & $3.58\times$ \\

        \textbf{+ FB}
        & 4 & \textbf{81.71} & $\mathbf{-2.67\pm0.65}$ & $2.08\times$ \\

        \bottomrule
    \end{tabular}
    \vspace{-0.8em}
\end{wraptable}
increasing accuracy degradation as more queries are superposed.
Margin-gated fallback substantially improves this tradeoff by selectively recomputing the
lowest-confidence 20\% of predictions. With adaptation and fallback, $K=2$
matches Plain accuracy ($+0.37\pm0.42$ pp) while providing $1.39\times$
speedup; $K=3$ achieves $1.78\times$ speedup with only a
$1.16\pm0.51$ pp average accuracy loss; and $K=4$ reaches $2.08\times$
speedup with a $2.67\pm0.65$ pp loss. These results establish $K$ as an
explicit deployment knob: larger $K$ amortizes more encoding computation,
while selective fallback provides a complementary mechanism for recovering
prediction fidelity.

\textbf{Complementary Efficiency Knobs:}
SupHDC is intended to complement, rather than replace, existing ways of
controlling HDC efficiency. Figure~\ref{fig:dimension_quantization}(a) varies
the random-feature dimension while keeping the SupHDC procedure fixed.
Larger representations consistently tolerate superposition better: at $K=4$,
for example, the average accuracy loss decreases from $8.31$ pp at $D=1{,}000$ to
$2.67$ pp at $D=10{,}000$. Thus, reducing $D$ and increasing $K$ are not equivalent
operations. The former changes the representation capacity and redundancy,
whereas the latter amortizes inference across multiple queries while retaining
the chosen representation dimension. The dependence on $D$ here reflects
finite-feature score fidelity rather than the input-dimensional $d$ dependence
of Proposition~\ref{prop:key_alignment}. Figure~\ref{fig:dimension_quantization}(b) similarly evaluates SupHDC under
different numerical precisions, always relative to a Plain model using the
same precision. SupHDC provides the same analytical savings of
$1.39\times$, $1.78\times$, and $2.08\times$ for $K=2,3,4$ across the
evaluated precision settings. Moreover, 8-bit and 4-bit configurations closely
track the full-precision accuracy behavior throughout this practical regime,
whereas more aggressive 1--2 bit quantization introduces larger degradation.
This shows that quantization and superposed inference are complementary:
SupHDC retains its accuracy--compute behavior under 4--8 bit numerical
quantization while reducing the number of encodings.

\textbf{Understanding the Superposition Regime:}
The theory separates two sources of distortion: cross-slot
interaction, governed by input dimension $d$ and superposition factor $K$;
and finite random-feature approximation, controlled by $D$.
For the final classifier, we measure the relative score distortion:
\begin{equation}
\small
    R_{\mathrm{int}}
    =
    \frac{\|\tilde{\mathbf{s}}-\rho\mathbf{s}\|_2}
         {\|\rho\mathbf{s}\|_2}
    \label{eq:relative_interference}
\end{equation}
where $\mathbf{s}$ and $\tilde{\mathbf{s}}$ denote the clean and superposed
score vectors. The normalization removes the effect of positive common
scaling $\rho$, which alone does not change class ordering.

\begin{figure}[t]
    \centering
    \includegraphics[
        width=1\linewidth
    ]{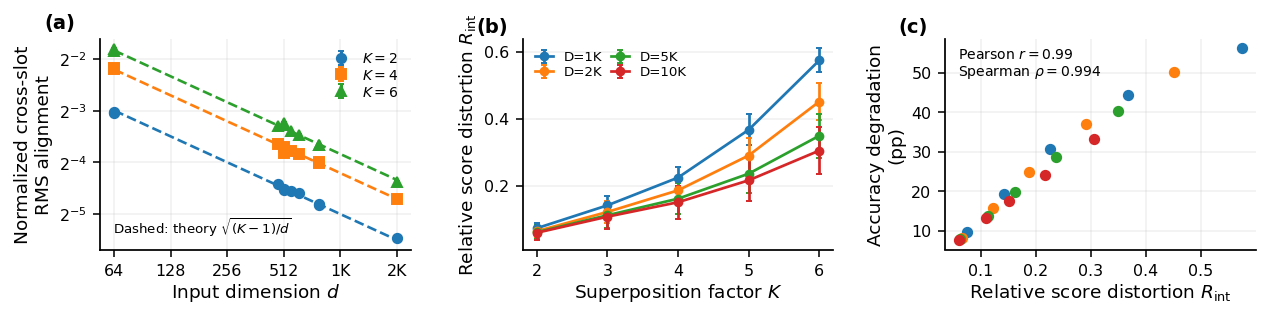}
    \vspace{-10mm}
    \caption{
    \textbf{From key separation to prediction accuracy.}
    (a) On the actual $\ell_2$-normalized benchmark inputs, cross-slot
    alignment closely follows the $\sqrt{(K-1)/d}$ dependence predicted by
    Proposition~\ref{prop:key_alignment}; dashed curves show the theoretical
    values without fitting.
    (b) During full SupHDC inference, relative score distortion increases with
    superposition factor $K$, while larger random-feature dimension $D$
    reduces the finite-feature contribution to this distortion.
    (c) Across the shown $(D,K)$ operating points, larger score distortion
    strongly corresponds to greater accuracy degradation.
    }
    \label{fig:theory_validation}
    \vspace{-5mm}
\end{figure}

Figure~\ref{fig:theory_validation}(a) directly validates the random-key
separation result of Proposition~\ref{prop:key_alignment}. For
$\ell_2$-normalized inputs, the predicted root mean square cross-slot alignment scales as
$\sqrt{(K-1)/d}$. Across the benchmark inputs and
$K\in\{2,4,6\}$, the measured values closely follow these parameter-free
predictions, with a mean empirical-to-theoretical ratio of $1.001$.
Thus, increasing $K$ raises the structural interference load, while larger
input dimension $d$ reduces coherent cross-slot alignment.
Panel (b) examines the full finite-dimensional classifier. The measured
relative score distortion increases systematically with $K$, while increasing
the random-feature dimension $D$ reduces the additional finite-feature
approximation error. Importantly, this role of $D$ is distinct from the
$1/\sqrt{d}$ key-separation effect in panel (a).
Finally, Fig.~\ref{fig:theory_validation}(c) connects the score analysis to
prediction quality. Across the practical $K=2$--$6$ operating points,
$R_{\mathrm{int}}$ strongly tracks accuracy degradation
(Pearson $r=0.99$, Spearman $\rho=0.994$). These results support the central
view of SupHDC: exact recovery of each individual hypervector is unnecessary;
prediction is preserved when class-dependent score distortion remains small
relative to the scaled clean evidence.

\textbf{Edge Deployment:}
We finally evaluate whether SupHDC's analytical compute savings translate to
wall-clock acceleration on resource-constrained hardware. We deploy the
full-precision $D=10{,}000$ models on a Raspberry Pi~5 and evaluate HAR,
ISOLET, CIFAR-10, and AG News for $K\in\{2,4\}$. SupHDC uses the same superposition-aware prototype adaptation as in the main
experiments, while the fallback variant additionally cleanly recomputes the
lowest-confidence 20\% of queries. All measurements use a single CPU thread. For each dataset and configuration, we time blocks of
\begin{wraptable}{r}{0.47\columnwidth}
    \vspace{-2em}
    \centering
    \small
    \setlength{\tabcolsep}{2.5pt}
    \caption{
    \textbf{Raspberry Pi~5 deployment.}
    $D=10{,}000$, full precision, one CPU thread.
    }
    \label{tab:pi}
    \vspace{0.3em}
    \begin{tabular}{lccc}
        \toprule
        Setting
        & Meas.
        & Analyt.
        & $\Delta$ Acc. \\
        & speedup
        & speedup
        & (pp) \\
        \midrule
        $K=2$, Adapt.
        & $1.79\times$
        & $1.90\times$
        & $-2.58$ \\

        $K=2$, Adapt. + FB
        & $1.19\times$
        & $1.38\times$
        & $-0.47$ \\

        $K=4$, Adapt.
        & $3.36\times$
        & $3.45\times$
        & $-6.66$ \\

        $K=4$, Adapt. + FB
        & $2.01\times$
        & $2.04\times$
        & $-2.26$ \\
        \bottomrule
    \end{tabular}
    \vspace{-0.8em}
\end{wraptable}
128 inference groups over 30 repetitions. Training and model loading are
excluded from timing. Measured speedup is computed from wall-clock latency
relative to the corresponding batched Plain HDC implementation and aggregated
across datasets using the geometric mean. Analytical speedup is aggregated in
the same way, while $\Delta$ Acc. reports the mean accuracy change relative to
paired Plain HDC.
Table~\ref{tab:pi} shows that the analytical savings translate into
substantial hardware acceleration. Without fallback, SupHDC reaches
$1.79\times$ measured speedup at $K=2$ and $3.36\times$ at $K=4$, closely
tracking analytical estimates of $1.90\times$ and $3.45\times$,
respectively. Selective fallback provides a direct accuracy--speed tradeoff:
at $K=4$, recomputing the lowest-confidence 20\% of queries reduces the mean
accuracy loss from $6.66$ to $2.26$ pp while retaining $2.01\times$ measured
speedup. At $K=2$, the same fallback reduces the loss to $0.47$ pp while
retaining $1.19\times$ acceleration.

\section{Conclusion}

We introduced \textbf{SupHDC}, a new inference paradigm for hyperdimensional computing that replaces repeated per-query encoding with shared superposed inference. By viewing projection-based HDC through the lens of kernel machines, we show that inference need not recover each representation exactly. It is sufficient to preserve the class evidence required for prediction. This perspective enables multiple independent queries to share one high-dimensional encoding computation, with random slot keys, superposition-aware adaptation, and selective fallback controlling the resulting interference. Across diverse datasets and edge deployment, SupHDC achieves around $2\times$ inference speedup with minimal accuracy loss. More broadly, SupHDC suggests that the redundancy of high-dimensional representations can serve not only as a source of robustness, but also as computational capacity for multiplexed inference, opening a new direction for HDC and kernel machines in which multiple predictions are carried through a shared representation computation.

\section*{Acknowledgment}
This work was supported in part by PRISM and CoCoSys, centers in JUMP 2.0, an SRC program sponsored by DARPA (SRC grant number - 2023-JU-3135). This work was also supported by NSF grants \#2003279, \#1911095, \#2112167, \#2052809, \#2112665, \#2120019, \#2211386.

%%%%%%%%%%%%%%%%%%%%%%%%%%%%%%%%%%%%%%%%%%%%%%%%%%%%%%%%%%%%
\bibliographystyle{iclr2027_conference}
\bibliography{iclr2027_conference}
%%%%%%%%%%%%%%%%%%%%%%%%%%%%%%%%%%%%%%%%%%%%%%%%%%%%%%%%%%%%

%%%%%%%%%%%%%%%%%%%%%%%%%%%%%%%%%%%%%%%%%%%%%%%%%%%%%%%%%%%%
\newpage
\appendix

The appendix here provides additional details for the submission titled: “Superposed Inference for Hyperdimensional Computing”. The appendix is organized as follows:

{\setstretch{2}
\begin{enumerate}[label=\Alph*.]
    \large
    \item \hyperref[app:notation]{\textbf{List of notation}}
    \item \hyperref[app:proofs]{\textbf{Proofs}}
    \item \hyperref[app:implementation]{\textbf{SupHDC Algorithm and Implementation Details}}
    \item \hyperref[app:experimental_protocol]{\textbf{Datasets, Preprocessing, and Experimental Protocol}}
    \item \hyperref[app:extended_results]{\textbf{Extended Accuracy--Compute Results}}
    \item \hyperref[app:ablations]{\textbf{Ablations of Adaptation and Selective Fallback}}
    \item \hyperref[app:cost_storage]{\textbf{Detailed Computational and Storage Analysis}}
    \item \hyperref[app:edge_results]{\textbf{Additional Edge Deployment Results}}
    \item \hyperref[app:discussion]{\textbf{Discussion and Limitations}}
\end{enumerate}
}
\newpage

\section{List of notations}
\label{app:notation}

We hereby provide a list of notations used in this paper:

\begin{table*}[h]
\caption{List of notations.}
\begin{center}
\small
\setlength{\tabcolsep}{6pt}
\begin{tabular}{p{8em} || p{33em}} 
 \toprule
 \toprule
 Symbol & Meaning \\
 \midrule

 \centering \(x \in \mathbb{R}^{d}\) 
 & Input query in the original feature space. \\

 \centering \(d\) 
 & Original input dimension. \\

 \centering \(D\) 
 & Hyperdimensional / random-feature representation dimension. \\

 \centering \(C\) 
 & Number of classes. \\

 \centering \(K\) 
 & Superposition factor; number of queries processed through one shared encoding. \\

 \centering \(W\in\mathbb{R}^{D\times d}\) 
 & Fixed random-feature projection matrix. \\

 \centering \(\phi_D(x)\) 
 & \(D\)-dimensional complex random-feature encoder. \\

 \centering \(P_c\) 
 & Class prototype in the base HDC classifier. \\

 \centering \(\alpha_{c,i}\) 
 & Learned contribution of training example \(x_i\) to the prototype of class \(c\). \\

 \centering \(A_k\) 
 & Random signed-permutation key assigned to slot \(k\). \\

 \centering \(x_{\mathrm{mix}}\) 
 & Superposed input,
 \(x_{\mathrm{mix}}=\sum_{k=1}^{K}A_kx_k\). \\

 \centering \(P_{k,c}\) 
 & Slot-specific prototype for slot \(k\) and class \(c\). \\

 \centering \(\tilde{s}_{k,c}\) 
 & Superposed class score for slot \(k\) and class \(c\). \\

 \centering \(q\) 
 & Fraction of low-confidence predictions selected for clean fallback. \\

 \centering \(\kappa(\cdot)\) 
 & Shift-invariant kernel expressed as a function of input displacement. \\

 \centering \(v_k\) 
 & Aggregate interference from all slots other than \(k\),
 \(v_k=\sum_{\ell\neq k}A_\ell x_\ell\). \\

 \centering \(\delta_{k,i}\) 
 & Keyed displacement between query \(x_k\) and training example \(x_i\),
 \(\delta_{k,i}=A_k(x_k-x_i)\). \\

 \centering \(r_{\kappa}(\delta,v)\) 
 & Kernel residual capturing the interaction between a keyed displacement and superposition interference. \\

 \centering \(\rho_k\) 
 & Class-independent scaling of the clean score induced by superposition. \\

 \centering \(\epsilon_{k,c}\) 
 & Class-dependent score distortion caused by superposition. \\

 \centering \(\gamma_k\) 
 & Clean classification margin for slot \(k\). \\

 \centering \(\sigma\) 
 & RBF kernel bandwidth. \\

 \centering \(R_{\mathrm{int}}\) 
 & Relative score-distortion metric used to quantify superposition interference. \\

 \centering \(E,\;R\) 
 & Cost of one high-dimensional encoding and one prototype readout, respectively. \\

 \bottomrule
 \bottomrule
\end{tabular}
\label{notation}
\end{center}
\end{table*}

\section{Proofs}
\label{app:proofs}

This section provides derivations and proofs for the theoretical results in
Section~\ref{sec:theory}. We use the same notation as in the main text. For
slot $k$, $u_k=A_kx_k, v_k=\sum_{\ell\neq k}A_\ell x_\ell$,
and for a keyed training example $A_kx_i$, $\delta_{k,i}=A_k(x_k-x_i)$.

\subsection{Derivation of the Score Decomposition}
\label{app:score_decomposition}

We first derive Eq.~\ref{eq:score_decomposition}. At the underlying
kernel level, write the clean class score for slot $k$ as
\begin{equation}
    s_{k,c}
    =
    \sum_i
    \beta_{k,c,i}
    \kappa(\delta_{k,i})
    \label{eq:app_clean_score}
\end{equation}
where prototype normalization is absorbed into the coefficients
$\beta_{k,c,i}$.

Under superposition, the query displacement becomes
$\delta_{k,i}+v_k$, giving
\begin{equation}
    \tilde{s}_{k,c}
    =
    \sum_i
    \beta_{k,c,i}
    \kappa(\delta_{k,i}+v_k)
    \label{eq:app_superposed_score}
\end{equation}
Define
\begin{equation}
r_\kappa(\delta,v)
=
\kappa(\delta+v)-\kappa(\delta)\kappa(v)
\end{equation}
Then
\begin{equation}
\kappa(\delta+v)
=
\kappa(v)\kappa(\delta)+r_\kappa(\delta,v)
\end{equation}
and substituting this identity into
Eq.~\ref{eq:app_superposed_score} gives
\begin{align}
    \tilde{s}_{k,c}
    &=
    \kappa(v_k)
    \sum_i \beta_{k,c,i}\kappa(\delta_{k,i})
    +
    \sum_i
    \beta_{k,c,i}
    r_\kappa(\delta_{k,i},v_k)
    \nonumber\\
    &=
    \rho_k s_{k,c}+\epsilon_{k,c}
\end{align}
where
\begin{equation}
\rho_k=\kappa(v_k),
\qquad
\epsilon_{k,c}
=
\sum_i
\beta_{k,c,i}
r_\kappa(\delta_{k,i},v_k)
\end{equation}
This is Eq.~\ref{eq:score_decomposition}. Importantly, $\rho_k$ is common
to all classes within slot $k$, whereas $\epsilon_{k,c}$ is class dependent.

\subsection{RBF Residual and Residual Bound}
\label{app:rbf_residual}

For the RBF kernel
\begin{equation}
\kappa(t)
=
\exp\!\left(
-\frac{\|t\|_2^2}{2\sigma^2}
\right)
\end{equation}
the identity
\begin{equation}
\|\delta+v\|_2^2
=
\|\delta\|_2^2+\|v\|_2^2+2\delta^\top v
\end{equation}
gives
\begin{equation}
\kappa(\delta+v)
=
\kappa(\delta)\kappa(v)
\exp\!\left(
-\frac{\delta^\top v}{\sigma^2}
\right)
\end{equation}
Therefore,
\begin{equation}
    r_\kappa(\delta,v)
    =
    \kappa(\delta)\kappa(v)
    \left[
    \exp\!\left(
    -\frac{\delta^\top v}{\sigma^2}
    \right)-1
    \right]
\end{equation}
which is Eq.~\ref{eq:rbf_residual}.

Using
\begin{equation}
|e^{-t}-1|
\leq
e^{|t|}|t|
\end{equation}
with $t=\delta^\top v/\sigma^2$ yields
\begin{equation}
    |r_\kappa(\delta,v)|
    \leq
    \kappa(\delta)\kappa(v)
    \exp\!\left(
    \frac{|\delta^\top v|}{\sigma^2}
    \right)
    \frac{|\delta^\top v|}{\sigma^2}
\end{equation}
which is Eq.~\ref{eq:rbf_residual_bound}. Thus, for the RBF kernel, the
superposition residual is controlled directly by the cross-slot alignment
$\delta^\top v$ relative to the kernel scale $\sigma^2$.

\subsection{Proof of Proposition~\ref{prop:key_alignment}}
\label{app:proof_key_alignment}

We use the following property of a uniformly random signed-permutation matrix
$A=S\Pi$. For fixed $p,q\in\mathbb{R}^d$,
\begin{equation}
    \mathbb{E}[p^\top Aq]=0,
    \qquad
    \mathbb{E}\!\left[(p^\top Aq)^2\right]
    =
    \frac{\|p\|_2^2\|q\|_2^2}{d}
    \label{eq:app_signed_perm_moment}
\end{equation}
The first identity follows from the independent zero-mean random signs.
For the second, conditioning on the permutation eliminates all cross terms
between different coordinates, and averaging over the uniform permutation
assigns each squared coordinate of $q$ equally often to every coordinate of
$p$.

Let
\begin{equation}
a=x_k-x_i,
\qquad
\delta=A_ka,
\qquad
v_k=\sum_{\ell\neq k}A_\ell x_\ell
\end{equation}
Then
\begin{equation}
    \delta^\top v_k
    =
    \sum_{\ell\neq k}
    a^\top A_k^\top A_\ell x_\ell
    \label{eq:app_alignment_sum}
\end{equation}

For each $\ell\neq k$, the relative transformation
$A_k^\top A_\ell$ has the distribution of a uniform signed-permutation
matrix. This remains true when one of the two slots is slot $1$, since
$A_1=I_d$ and the transpose of a uniform signed-permutation matrix has the
same distribution. Hence, by Eq.~\ref{eq:app_signed_perm_moment},
\begin{equation}
\mathbb{E}
\left[
a^\top A_k^\top A_\ell x_\ell
\right]
=0
\end{equation}
and therefore
\begin{equation}
\mathbb{E}[\delta^\top v_k]=0
\end{equation}

For the second moment, the contribution of each interference slot is
\begin{equation}
\mathbb{E}
\left[
\left(
a^\top A_k^\top A_\ell x_\ell
\right)^2
\right]
=
\frac{\|a\|_2^2\|x_\ell\|_2^2}{d}
\end{equation}
The cross terms between distinct interference slots vanish. For $k=1$, this
follows from independence of the random keys. For $k\geq2$, conditioning on
$A_k$ makes the terms associated with distinct random $A_\ell$ zero-mean and
independent; if one of the terms corresponds to $A_1=I_d$, the other term
still has zero conditional mean.

Consequently,
\begin{equation}
    \mathbb{E}
    \left[
    (\delta^\top v_k)^2
    \right]
    =
    \frac{\|a\|_2^2}{d}
    \sum_{\ell\neq k}\|x_\ell\|_2^2
\end{equation}
which proves Eq.~\ref{eq:key_alignment_moment}.

Finally, applying Markov's inequality to the nonnegative random variable
$(\delta^\top v_k)^2$ gives, for any $\eta\in(0,1)$,
\begin{equation}
\Pr\!\left(
|\delta^\top v_k|
\geq
\sqrt{
\frac{
\mathbb{E}[(\delta^\top v_k)^2]
}{\eta}}
\right)
\leq \eta
\end{equation}
Substituting the second moment above yields, with probability at least
$1-\eta$,
\begin{equation}
|\delta^\top v_k|
\leq
\frac{\|a\|_2}{\sqrt d}
\sqrt{
\frac{
\sum_{\ell\neq k}\|x_\ell\|_2^2
}{\eta}}
\end{equation}
which is Eq.~\ref{eq:key_alignment_prob}.

\begin{flushright}
$\blacksquare$
\end{flushright}

\subsection{Proof of Proposition~\ref{prop:prediction_preservation}}
\label{app:proof_prediction_preservation}

Let $y_k$ be the unique clean prediction with margin
\begin{equation}
\gamma_k
=
s_{k,y_k}-\max_{c\neq y_k}s_{k,c}>0.
\end{equation}
For any competing class $c\neq y_k$
\begin{align}
    \tilde{s}_{k,y_k}-\tilde{s}_{k,c}
    &=
    \rho_k
    \left(
    s_{k,y_k}-s_{k,c}
    \right)
    +
    \epsilon_{k,y_k}-\epsilon_{k,c}
    \nonumber\\
    &\geq
    \rho_k\gamma_k
    -
    2\|\epsilon_k\|_\infty
\end{align}
Therefore, if
\begin{equation}
2\|\epsilon_k\|_\infty<\rho_k\gamma_k
\end{equation}
then
$\tilde{s}_{k,y_k}>\tilde{s}_{k,c}$ for every $c\neq y_k$, and hence
\begin{equation}
\arg\max_c\tilde{s}_{k,c}=y_k
\end{equation}

\begin{flushright}
$\blacksquare$
\end{flushright}

Notice that $\rho_k$ need not be close to one. When $\rho_k>0$, common
scaling alone preserves the class ordering. Its effect in the proposition is
to scale the clean margin available to tolerate class-dependent distortion.

\subsection{Finite-$D$ Random-Feature Approximation}
\label{app:finiteD}

The next analysis concerns the underlying kernel approximation. With standard
i.i.d. random Fourier features,
\begin{equation}
\phi_D(x)
=
\frac{1}{\sqrt D}
\left[
e^{i\omega_1^\top x}
\ldots,
e^{i\omega_D^\top x}
\right]
\end{equation}
the empirical kernel is
\begin{equation}
    \hat{k}_D(x,x')
    =
    \operatorname{Re}
    \langle
    \phi_D(x),\phi_D(x')
    \rangle
    =
    \frac{1}{D}
    \sum_{j=1}^{D}
    \cos\!\left(
    \omega_j^\top(x-x')
    \right)
    \label{eq:app_empirical_kernel}
\end{equation}
By the random Fourier feature construction,
\begin{equation}
\mathbb{E}[\hat{k}_D(x,x')]
=
k(x,x')
\end{equation}
for the corresponding shift-invariant kernel
\citep{rahimi2007random}. Since each summand in
Eq.~\ref{eq:app_empirical_kernel} lies in $[-1,1]$, Hoeffding's inequality
gives, for fixed $x,x'$,
\begin{equation}
    \Pr\!\left(
    |\hat{k}_D(x,x')-k(x,x')|
    \geq\varepsilon
    \right)
    \leq
    2\exp\!\left(
    -\frac{D\varepsilon^2}{2}
    \right)
    \label{eq:app_rff_concentration}
\end{equation}
Thus the pointwise approximation error is
$O(D^{-1/2})$, consistent with standard RFF analysis
\citep{sutherland2015error}.

Accordingly, the finite-dimensional superposed score can be written as
\begin{equation}
\tilde{s}^{(D)}_{k,c}
=
\tilde{s}_{k,c}
+
\epsilon^{\mathrm{RFF}}_{k,c}
\end{equation}
Combining this with Eq.~\ref{eq:score_decomposition} gives
\begin{equation}
\tilde{s}^{(D)}_{k,c}
=
\rho_k s_{k,c}
+
\epsilon^{\mathrm{sup}}_{k,c}
+
\epsilon^{\mathrm{RFF}}_{k,c}
\end{equation}
where
$\epsilon^{\mathrm{sup}}_{k,c}=\epsilon_{k,c}$ denotes structural
superposition interference and
$\epsilon^{\mathrm{RFF}}_{k,c}$ denotes finite-feature approximation error.
This is Eq.~\ref{eq:finiteD}.

The concentration result above is stated for standard i.i.d. RFF, as in the
analysis of Section~\ref{sec:theory}. Our implementation uses orthogonal random features (ORF)~\citep{yu2016orthogonal}, which reduces kernel-estimation variance
relative to standard i.i.d. RFF, thereby improving
finite-$D$ approximation accuracy without changing the $O(D^{-1/2})$
convergence order \citep{yu2016orthogonal}.

\section{SupHDC Algorithm and Implementation Details}
\label{app:implementation}

\subsection{Base HDC Model Training}

SupHDC starts from a trained projection-based HDC classifier. Our
implementation uses the complex random-feature encoder
\begin{equation}
    \phi_D(x)
    =
    \frac{1}{\sqrt{D}}\exp(iWx)
\end{equation}
where $W\in\mathbb{R}^{D\times d}$ is constructed using orthogonal random
features (ORF) and is sampled once and kept fixed thereafter. Each ORF block
is obtained by orthogonalizing a Gaussian matrix with QR decomposition\citep{yu2016orthogonal}. We use kernel bandwidth $\sigma=1$.

The base classifier follows the common prototype-based HDC training
procedure: inputs are first encoded into hypervectors, examples from the same
class are bundled into a class prototype, and the prototypes are
refined using prediction errors. This general encode--bundle--compare
procedure covers a broad class of existing HDC works, including but not limited to AdaptHD and OnlineHD, OpenHD and DPQ-HD\citep{hernandez2021onlinehd,imani2019adapthd,pandey2025dpq,kang2022openhd}. Thus, SupHDC does not depend on a
special-purpose base classifier and is generally applicable to the HDC community.

Algorithm~\ref{alg:base_hdc} summarizes the base training used in our
experiments. We initialize each class prototype by summing the encoded
training examples from that class. During refinement, examples are shuffled
and processed in minibatches. Classification uses the normalized prototypes.
For every misclassified example, the true-class prototype is moved toward
the encoded example and the incorrectly predicted prototype is moved away
from it, with the update magnitude determined by their current scores.

\begin{algorithm}[H]
\caption{Base RFF-HDC Training}
\label{alg:base_hdc}
\begin{algorithmic}[1]
\REQUIRE Training data $\{(x_i,y_i)\}_{i=1}^{n}$,
         fixed encoder $\phi_D(\cdot)$,
         learning rate $\eta$,
         number of training epochs $E$
\ENSURE Class prototypes $\{P_c\}_{c=1}^{C}$ and sample weights
        $\{\alpha_{c,i}\}$

\STATE Initialize $P_c \leftarrow
       \sum_{i:y_i=c}\phi_D(x_i)$ for each class $c$
\STATE Initialize
       $\alpha_{c,i}\leftarrow\mathbf{1}[y_i=c]$

\FOR{$e=1,\ldots,E$}
    \STATE Randomly shuffle the training examples
    \STATE Normalize prototypes:
           $\bar P_c\leftarrow P_c/\|P_c\|_2$
    \FOR{each training minibatch}
        \STATE Compute
        $s_c(x_i)=\operatorname{Re}
        \langle\phi_D(x_i),\bar P_c\rangle$
        for all examples and classes
        \STATE $\hat y_i\leftarrow\arg\max_c s_c(x_i)$
        \FOR{each example with $\hat y_i\neq y_i$}
            \STATE $w_i^{+}\leftarrow
                   \eta(1-s_{y_i}(x_i))$
            \STATE $w_i^{-}\leftarrow
                   \eta s_{\hat y_i}(x_i)$
            \STATE $P_{y_i}\leftarrow
                   P_{y_i}+w_i^{+}\phi_D(x_i)$
            \STATE $P_{\hat y_i}\leftarrow
                   P_{\hat y_i}-w_i^{-}\phi_D(x_i)$
            \STATE $\alpha_{y_i,i}\leftarrow
                   \alpha_{y_i,i}+w_i^{+}$
            \STATE $\alpha_{\hat y_i,i}\leftarrow
                   \alpha_{\hat y_i,i}-w_i^{-}$
        \ENDFOR
        \STATE Refresh normalized prototypes
    \ENDFOR
\ENDFOR
\STATE $P_c\leftarrow P_c/\|P_c\|_2$
\RETURN $\{P_c\}_{c=1}^{C}$ and $\{\alpha_{c,i}\}$
\end{algorithmic}
\end{algorithm}

The resulting unnormalized prototype has the equivalent form
\begin{equation}
    P_c^{\mathrm{sum}}
    =
    \sum_{i=1}^{n}
    \alpha_{c,i}\phi_D(x_i).
    \label{eq:app_base_weighted_prototype}
\end{equation}
SupHDC retains these learned sample weights when constructing the
slot-specific prototype banks under different slot keys. The base projection
matrix $W$ is never modified by SupHDC.

\subsection{Full SupHDC Algorithm}

SupHDC is applied on top of the trained base HDC model. For a superposition factor $K$, we first
construct $K$ fixed slot keys. The first slot uses the identity key,
$A_1=I_d$, while each remaining key is an independently sampled signed
permutation, $A_k=S_k\Pi_k$. The keys are sampled once and then kept fixed
throughout adaptation and inference.

For each slot, we construct a clean prototype bank by re-encoding the
training examples under that slot key while reusing the sample weights
$\alpha_{c,i}$ learned by the base model:
\begin{equation}
    P^{\mathrm{clean}}_{k,c}
    =
    \frac{
        \sum_i \alpha_{c,i}\phi_D(A_kx_i)
    }{
        \left\|
        \sum_i \alpha_{c,i}\phi_D(A_kx_i)
        \right\|_2
    }
    \label{eq:app_clean_slot_proto}
\end{equation}
The clean bank is retained unchanged for selective fallback. A separate
working copy is initialized from these normalized prototypes and adapted
using superposed training groups.

During superposition-aware adaptation, groups of $K$ training examples are
sampled randomly with replacement. The $K$ keyed inputs are summed before
encoding, so each group requires only one call to $\phi_D$. All $K$ slots
are scored from this shared representation. If a slot is misclassified,
only its true-class and predicted-class prototypes are updated using the
score-dependent correction described in the main text. The working
prototypes are then renormalized. The encoder, slot keys, and clean
prototype bank remain fixed.

\begin{algorithm}[h]
\caption{SupHDC Adaptation and Inference}
\label{alg:suphdc_full}
\begin{algorithmic}[1]
\REQUIRE Trained base model, superposition factor $K$,
         adaptation epochs $E$, fallback fraction $q$

\STATE Sample fixed slot keys $\{A_k\}_{k=1}^{K}$ with $A_1=I_d$
\STATE Build clean slot prototypes
       $\{P^{\mathrm{clean}}_{k,c}\}$ using the learned
       base-model weights $\alpha_{c,i}$
\STATE Initialize working prototypes
       $P_{k,c}\leftarrow P^{\mathrm{clean}}_{k,c}$

\FOR{$e=1,\ldots,E$}
    \FOR{each sampled group of $K$ training examples}
        \STATE Key and sum the $K$ inputs, then encode the mixture once
        \STATE Score all $K$ slots using their working prototypes
        \STATE Update the true and predicted prototypes of misclassified slots
        \STATE Normalize the working prototypes
    \ENDFOR
\ENDFOR

\STATE \textbf{Inference:} key and sum each group of $K$ queries
\STATE Encode each mixture once and obtain all $K$ slot predictions

\IF{$q>0$}
    \STATE Rank valid predictions by their top-1--top-2 score margin
    \STATE Select the lowest-margin fraction $q$
    \STATE Individually re-encode selected queries under their slot keys
    \STATE Replace their predictions using the clean prototype bank
\ENDIF

\RETURN One prediction for each input query
\end{algorithmic}
\end{algorithm}

The selective fallback is a clean recomputation rather than an interference
cancellation procedure. For an inference group-batch containing $N_b$ valid
slot predictions, we select exactly $\lceil qN_b\rceil$ predictions with the
smallest top-1--top-2 score margins. Only these queries are individually
re-encoded and rescored using $P^{\mathrm{clean}}_{k,c}$; all other
predictions retain the result of the shared superposed encoding.

\textbf{Quantization:}
For a $b$-bit setting, we apply the same bit width to both the projection
matrix $W$ and the encoded phase. For $b\geq2$, each row of $W$ is quantized
independently using uniform asymmetric min--max quantization with
$2^b$ levels. For $b=1$, we instead use a symmetric norm-preserving binary
quantizer,
\begin{equation}
    W_{j,:}^{(q)}
    =
    \frac{\|W_{j,:}\|_2}{\sqrt d}
    \operatorname{sign}(W_{j,:})
\end{equation}
which preserves the sampled ORF row norm. Phase quantization maps each
projected angle to its nearest point among $2^b$ uniformly spaced phases on
$[0,2\pi)$. Full precision corresponds to $b=0$, with neither weight nor
phase quantization. The low-bit PyTorch experiments simulate these numerical
quantization levels; the stored integer codes are not physically bit-packed
unless a dedicated low-bit kernel is used.

\subsection{Hyperparameters}

Table~\ref{tab:suphdc_hyperparameters} summarizes the hyperparameters used
in the main experiments. Unless otherwise stated, the same settings are used
for all datasets. For each fixed combination of dataset, $D$, precision, and
random seed, the same trained base model is reused across Plain HDC and all
values of $K$. This provides paired comparisons in which the projection
matrix, base-model training, and test ordering are shared.

\begin{table}[h]
\centering
\small
\caption{Hyperparameters used in the main SupHDC experiments.}
\label{tab:suphdc_hyperparameters}
\begin{tabular}{lll}
\toprule
\textbf{Parameter} & \textbf{Setting} & \textbf{Description} \\
\midrule

Kernel bandwidth $\sigma$
    & $1$
    & RBF/ORF bandwidth \\

Hypervector dimension $D$
    & $\{1000,2000,5000,10000\}$
    & Random-feature dimension \\

Superposition factor $K$
    & $\{2,\ldots,10\}$
    & Queries encoded jointly \\

Base training epochs
    & $20$
    & Prototype refinement epochs \\

Base learning rate $\eta$
    & $1.0$
    & Error-driven prototype update \\

Adaptation epochs
    & $100$
    & Superposition-aware refinement \\

Adaptation learning rate
    & $1.0$
    & Same as base training \\

Fallback fraction $q$
    & $0.20$
    & Fraction selectively recomputed \\

Random seeds
    & $\{0,1,2,3,4\}$
    & Five paired runs \\

Precision
    & FP, 8, 4, 2, 1 bit
    & Joint $W$/phase quantization \\

\midrule
Base training batch size
    & $64$
    & Training samples per minibatch \\

Clean prototype batch size
    & $128$
    & Keyed prototype construction \\

Adaptation group batch size
    & $64$
    & Superposed groups per update batch \\

Inference group batch size
    & $1024$
    & Superposed groups per batch \\

\bottomrule
\end{tabular}
\end{table}

For the quantized experiments, the same bit width is applied jointly to the
projection matrix $W$ and the encoded phase. Full precision corresponds to
no weight or phase quantization. The low-bit settings in our PyTorch
implementation simulate the corresponding numerical precision; they should
not be interpreted as physically bit-packed storage unless a dedicated
low-bit kernel is used.

All random-feature projections and slot keys are sampled once for a given
run and remain fixed during base training, SupHDC adaptation, and inference.
SupHDC therefore adapts only the slot-specific prototype banks.

\subsection{Edge Experiment Details}

We additionally evaluate SupHDC on a Raspberry Pi 5 to verify that the
analytical compute reduction translates into wall-clock speedup on a
resource-constrained CPU platform. We use HAR, ISOLET, CIFAR-10, and
AG News with $D=10{,}000$, full-precision inference, and
$K\in\{2,4\}$. The evaluated methods are batched Plain HDC, adapted SupHDC,
and adapted SupHDC with selective clean fallback.

All models are trained and adapted before deployment. In particular, the
base model and SupHDC prototype banks are prepared offline using the same
training procedure as in the main experiments, with 100 epochs of
superposition-aware adaptation. The projection matrix, class prototypes,
slot keys, and test data are then stored in portable inference bundles and
transferred to the Raspberry Pi. Model training, bundle loading, and accuracy
evaluation are excluded from the timed region.

To make the comparison fair, both Plain HDC and SupHDC process the same
blocks of queries. Each timing block contains 128 groups, corresponding to
$128K$ individual queries. Plain HDC encodes all $K$ queries independently,
whereas SupHDC performs one shared encoding per group. For the fallback
variant, the selected low-margin queries are re-encoded in a single batched
clean projection before slot-specific classification.

All timing experiments use exactly one CPU thread. We perform five warm-up
runs followed by 30 measured repetitions for each configuration. The order
of the evaluated methods is rotated across repetitions so that thermal drift
does not systematically favor one method. We report the median latency over
the measured repetitions, and compute wall-clock speedup as
\begin{equation}
    S_{\mathrm{measured}}
    =
    \frac{
        T_{\mathrm{Plain}}
    }{
        T_{\mathrm{SupHDC}}
    }
\end{equation}
where both latencies are measured on the same query-block size. We also control for thermal effects between configurations. Before timing a
new dataset--$K$ pair, the benchmark waits until the CPU temperature falls
below $70^\circ$C and the current Raspberry Pi throttling flags are clear.
CPU temperature and throttling state are recorded before and after each
timed run.

The benchmark implementation evaluates fallback fractions
$q\in\{0.05,0.10,0.20\}$ without retraining the model. The main-paper
results use the $q=0.20$ operating point unless otherwise stated.

\section{Datasets, Preprocessing, and Experimental Protocol}
\label{app:experimental_protocol}

\subsection{Datasets and Input Representations}

We evaluate SupHDC on 10 classification datasets spanning sensing, speech,
vision, language, and manufacturing. Table~\ref{tab:datasets} summarizes the
final input representations provided to the HDC encoder. The reported sample
counts correspond to the actual feature matrices used in our experiments
after dataset preparation.

\begin{table*}[h]
\centering
\small
\caption{Datasets and input representations used in the SupHDC experiments.
$N_{\mathrm{tr}}$ and $N_{\mathrm{te}}$ denote the number of training and
test examples after preprocessing, $d$ is the input dimension, and $C$ is
the number of classes.}
\vspace{2mm}
\label{tab:datasets}
\begin{tabular}{@{}llrrrrp{4cm}@{}}
\toprule
\textbf{Dataset} &
\textbf{Domain} &
$\mathbf{N_{\mathrm{tr}}}$ &
$\mathbf{N_{\mathrm{te}}}$ &
$\mathbf{d}$ &
$\mathbf{C}$ &
\textbf{Input representation} \\
\midrule
HAR
    & Activity
    & 7,352 & 2,947 & 561 & 6
    & Activity feature vector \\

ISOLET
    & Speech
    & 6,238 & 1,559 & 617 & 26
    & Acoustic feature vector \\

CIFAR-10
    & Vision
    & 50,000 & 10,000 & 2,048 & 10
    & ResNet-50 feature \\

Airline Sentiment
    & Text
    & 9,228 & 2,307 & 64 & 2
    & Word2Vec feature \\

AG News
    & Text
    & 120,000 & 7,600 & 64 & 4
    & Word2Vec feature \\

IMDB
    & Text
    & 25,000 & 25,000 & 512 & 2
    & Word2Vec feature \\

20 Newsgroups
    & Text
    & 10,998 & 7,305 & 512 & 20
    & Word2Vec feature \\

SECOM
    & Manufacturing
    & 1,253 & 314 & 474 & 2
    & Process measurement vector \\

MNIST
    & Vision
    & 60,000 & 10,000 & 784 & 10
    & Flattened pixel vector \\

Fashion-MNIST
    & Vision
    & 60,000 & 10,000 & 784 & 10
    & Flattened pixel vector \\
\bottomrule
\end{tabular}
\end{table*}

For CIFAR-10, we use the output of an ImageNet-pretrained ResNet-50 with its
classification head removed, giving a 2,048-dimensional image
representation. For MNIST and Fashion-MNIST, the $28\times28$ images are
flattened directly into 784-dimensional vectors. HAR, ISOLET, and SECOM use
their provided tabular feature measurements.

For the four text datasets, each document is first tokenized and represented
using Word2Vec embeddings. We add a sinusoidal positional encoding to each
token embedding before mean pooling across the document, providing a
fixed-dimensional representation that retains simple word-order information.
Airline Sentiment and AG News use 64-dimensional embeddings, while IMDB and
20 Newsgroups use 512-dimensional embeddings. For Airline Sentiment, we keep
only positive and negative examples, resulting in a binary classification
task. 

For dataset-specific preprocessing, tabular and image-vector inputs are
standardized using training set statistics, while CIFAR-10 uses features extracted from
an ImageNet-pretrained ResNet-50 and the text datasets use the Word2Vec-based
representations described above. All final input vectors are
$\ell_2$-normalized before HDC encoding. This provides a consistent input
scale across datasets and matches the setting considered in our theoretical
analysis.

\subsection{Experimental Protocol and Result Aggregation}

The main evaluation considers hypervector dimensions
$D\in\{1{,}000,2{,}000,5{,}000,10{,}000\}$, superposition factors
$K\in\{2,\ldots,10\}$, and numerical precisions
$\{\mathrm{FP},8,4,2,1\}$ bits. For each configuration, we evaluate five
paired random seeds, $\{0,1,2,3,4\}$. We compare Plain RFF-HDC with four
SupHDC variants: direct superposed inference, superposed inference with
selective fallback, superposition-aware adaptation, and adaptation combined
with selective fallback. Accuracy is the primary predictive metric.

For a fixed dataset, $D$, precision, and random seed, a single trained base
model is shared by Plain HDC and all SupHDC configurations. In particular,
the same random-feature projection, base-model training, and test ordering
are reused across all values of $K$. This paired design ensures that changes
relative to Plain primarily reflect the effect of superposition rather than
independent model initialization. Test examples are deterministically shuffled
for each seed before being grouped into sets of $K$ queries.

For each dataset and operating point, results are first averaged over the
five random seeds. When reporting results across datasets, each dataset
contributes equally through its dataset-level mean. Accuracy changes are
reported in percentage points relative to the corresponding paired Plain
model, and uncertainty is reported as the standard error of the mean (SEM)
across dataset-level means. Repeated inference timings are used only to obtain
stable runtime measurements and are not treated as independent predictive
trials.

\section{Extended Accuracy--Compute Results}
\label{app:extended_results}

The main text focuses on the practical operating regime of SupHDC, with
$D=10{,}000$, full-precision inference, and superposition factors
$K\in\{2,3,4\}$. Here, we expand this evaluation to the complete
full-precision grid, covering $K=2,\ldots,10$, all four SupHDC variants,
and representation dimensions
$D\in\{1{,}000,2{,}000,5{,}000,10{,}000\}$.
As in the main experiments, results are first averaged across the five
paired random seeds within each dataset, after which each dataset
contributes equally to the reported cross-dataset mean. Accuracy changes
are measured in percentage points relative to the corresponding paired
Plain HDC model.

\begin{figure}[h]
    \centering
    \includegraphics[width=0.7\linewidth]
    {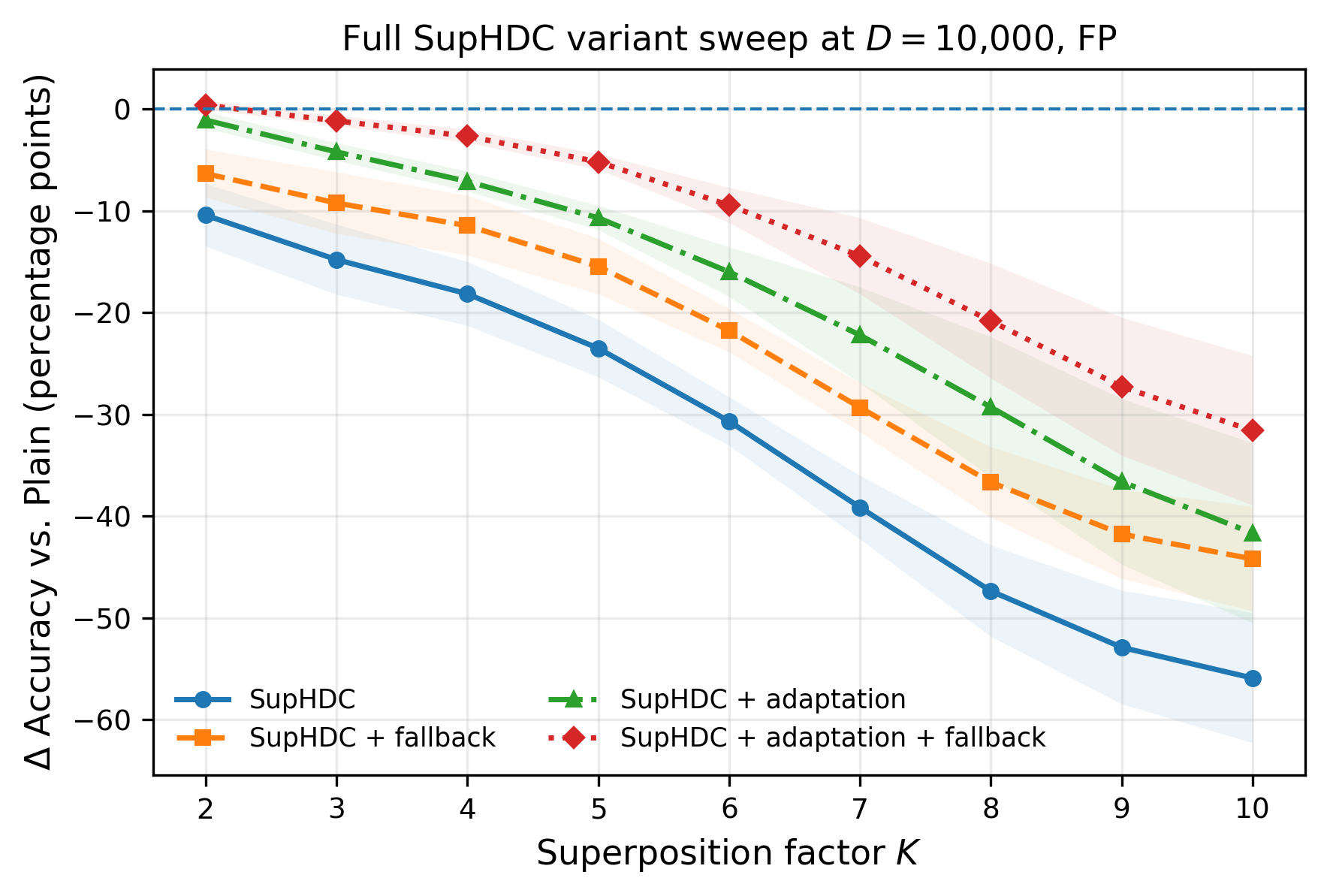}
    \caption{
    \textbf{Extended SupHDC variant sweep at $D=10{,}000$ and full
    precision.}
    Accuracy change relative to paired Plain HDC is shown for
    $K=2,\ldots,10$. Direct superposition degrades progressively as the
    interference load increases. Superposition-aware adaptation
    substantially improves accuracy without adding inference-time
    encoding operations, while selective fallback further recovers
    prediction fidelity by cleanly recomputing low-confidence queries.
    Shaded regions denote SEM across dataset-level means.
    }
    \label{fig:app_variant_sweep}
\end{figure}

\textbf{Behavior across the full superposition range:}
Figure~\ref{fig:app_variant_sweep} extends the $D=10{,}000$,
full-precision comparison to $K=2,\ldots,10$ for all four variants.
A consistent pattern emerges across the entire range. Direct
superposition incurs progressively larger accuracy degradation as more
queries share one encoding, reflecting the increasing interference load
at larger $K$. Selective fallback improves this behavior by cleanly
recomputing low-confidence predictions, while superposition-aware
adaptation provides a substantially larger improvement without adding
inference-time encoding operations. Combining adaptation and fallback
gives the strongest accuracy preservation throughout the sweep.

The extended range also clarifies why the main paper emphasizes
moderate superposition factors. At $K=2$, adaptation with fallback
matches the Plain HDC accuracy on average while still reducing inference
computation; at $K=3$ and $K=4$, it retains favorable tradeoffs with
only modest average degradation. Beyond this regime, the accuracy loss
grows increasingly rapidly. Thus, larger $K$ values are useful for
characterizing the stress regime of superposed inference, but do not
necessarily correspond to desirable deployment points.

\begin{figure}[h]
    \centering
    \includegraphics[width=0.7\linewidth]
    {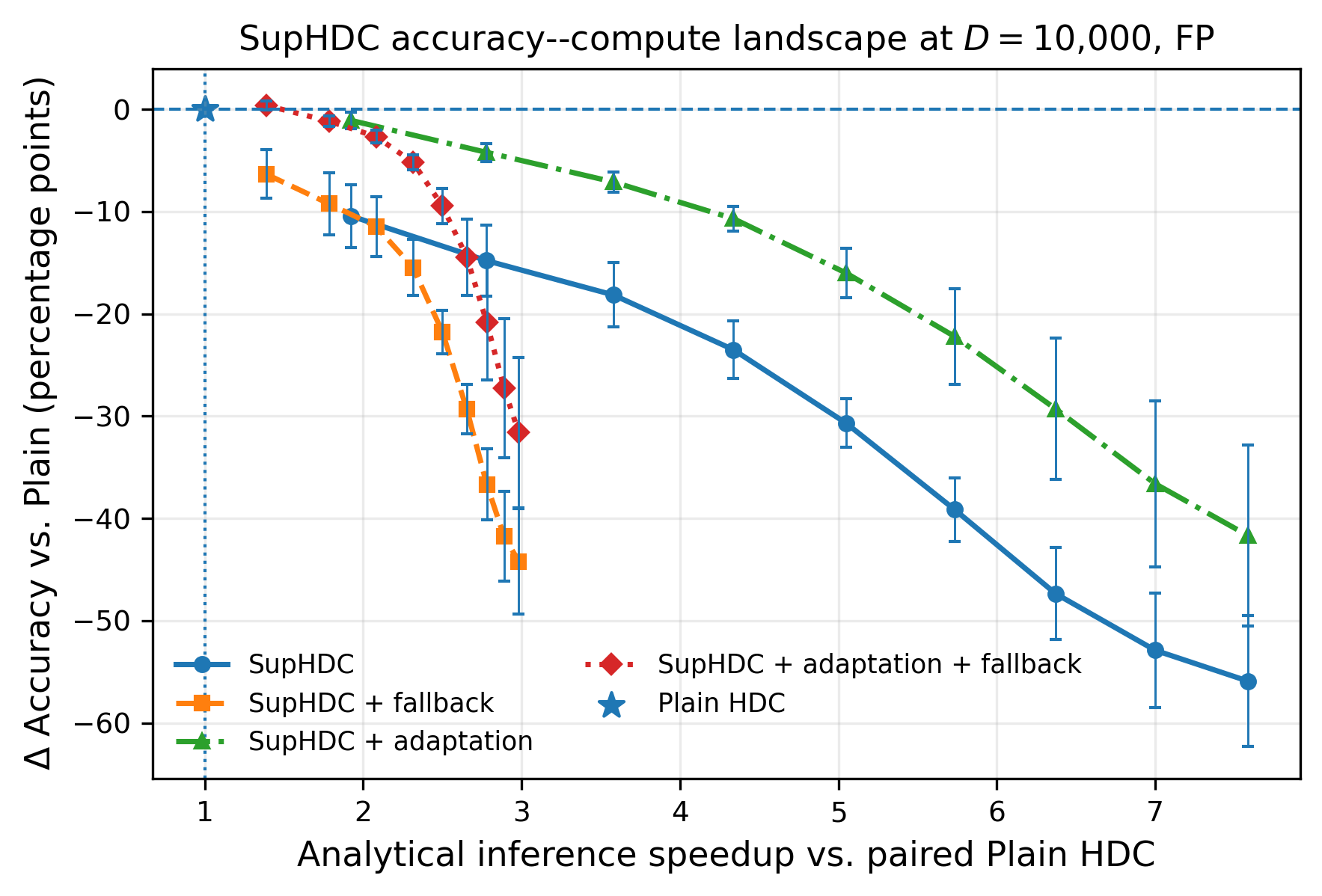}
    \caption{
    \textbf{Extended accuracy--compute landscape at $D=10{,}000$ and
    full precision.}
    Each point corresponds to a superposition factor
    $K\in\{2,\ldots,10\}$. Adaptation improves accuracy at essentially
    unchanged inference cost, whereas fallback deliberately trades part
    of the analytical speedup for clean recomputation of uncertain
    queries. The Plain HDC operating point is shown at $(1\times,0)$.
    Error bars denote SEM across dataset-level accuracy changes.
    }
    \label{fig:app_accuracy_compute}
\end{figure}

\textbf{Extended accuracy--compute landscape:}
Figure~\ref{fig:app_accuracy_compute} presents the same configurations
in accuracy--compute space. This view makes the distinct roles of the
two interference-mitigation mechanisms particularly clear.
Superposition-aware adaptation primarily moves an operating point
upward, improving predictive accuracy while leaving the inference
computation nearly unchanged. In contrast, selective fallback moves an
operating point upward and toward lower speedup, explicitly exchanging
part of the computational saving for improved prediction fidelity.
Their combination therefore provides a family of operating points
between clean inference and aggressive superposition.

The curves also reveal diminishing returns from very large $K$.
Although increasing $K$ continues to amortize the shared projection over
more queries, the incremental increase in analytical speedup becomes
smaller, whereas the accuracy degradation grows substantially. The
useful operating region is therefore determined not by maximizing $K$,
but by choosing a superposition factor for which the saved encoding
computation remains large relative to the induced score distortion.
This behavior is consistent with the practical $K=2$--$4$ points
emphasized in the main text.

\begin{figure}[h]
    \centering
    \includegraphics[width=0.7\linewidth]
    {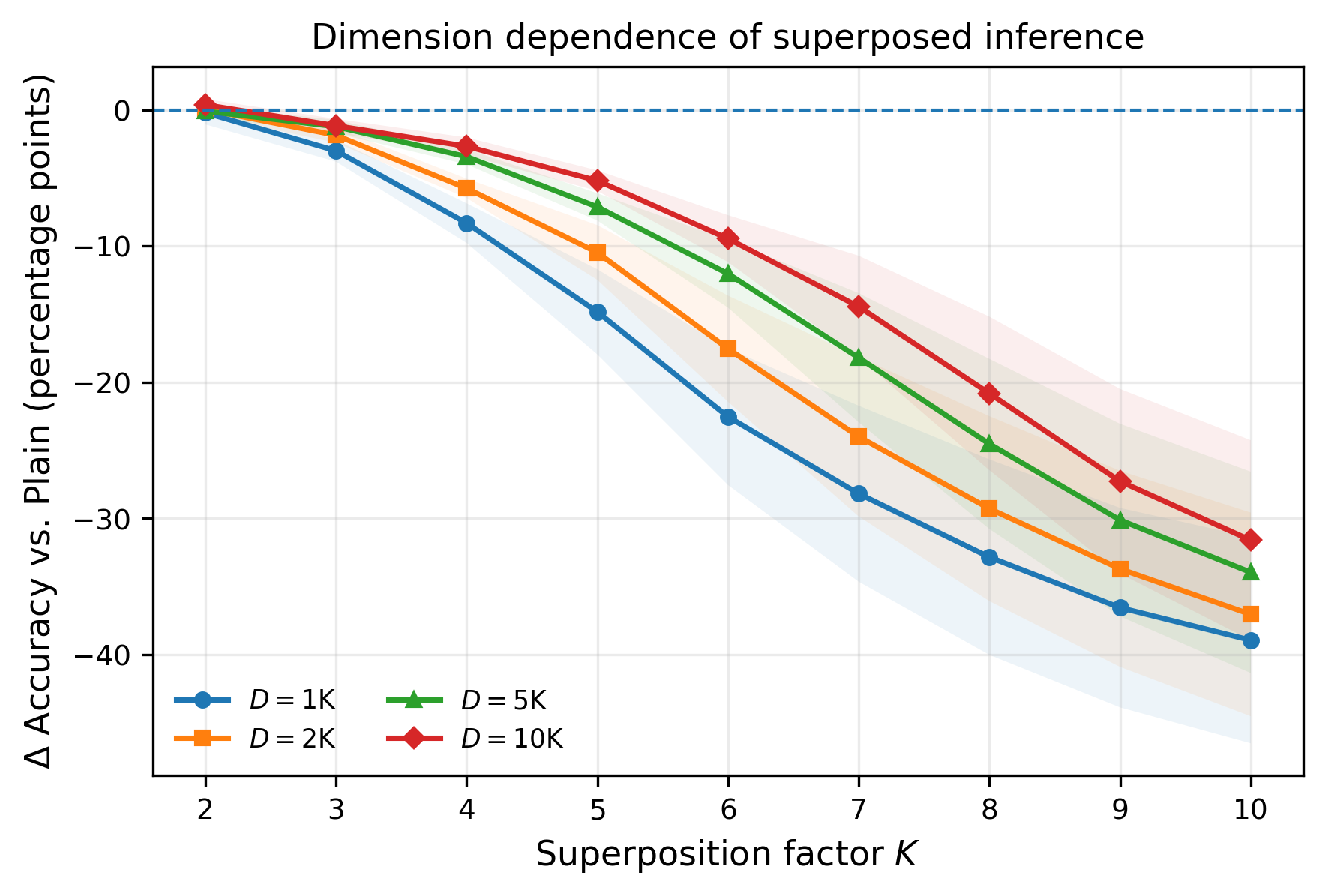}
    \caption{
    \textbf{Dimension dependence of superposed inference.}
    SupHDC remains effective across
    $D\in\{1{,}000,2{,}000,5{,}000,10{,}000\}$, while larger
    random-feature representations tolerate stronger superposition and
    therefore support more aggressive $K$.
    Accuracy change is shown for SupHDC with superposition-aware adaptation
    and $20\%$ selective fallback over $K=2,\ldots,10$ at full precision.
    Shaded regions denote SEM across dataset-level means.
    }
    \label{fig:app_dimension_sweep}
\end{figure}

\textbf{Effect of representation dimension:}
Figure~\ref{fig:app_dimension_sweep} isolates the effect of the
random-feature dimension $D$ using the final SupHDC configuration with
superposition-aware adaptation and $20\%$ selective fallback.
Increasing $D$ consistently improves tolerance to superposition, and
the difference becomes more pronounced as $K$ grows. This trend
continues into the higher-$K$ stress regime. Importantly, SupHDC is not restricted to the largest representation dimension: across all evaluated $D\in\{1K,2K,5K,10K\}$, moderate superposition factors provide useful accuracy--compute tradeoffs, while larger $D$ primarily extends the range of $K$ that can be tolerated.

This result reinforces the distinction between reducing representation
dimension and increasing the superposition factor. Reducing $D$ changes
the fidelity and redundancy of the random-feature representation itself,
whereas increasing $K$ amortizes inference computation while retaining
the selected representation dimension. Consequently, a high-dimensional
representation can provide additional headroom for shared inference:
when stronger finite-feature fidelity is retained, the classifier can
tolerate more superposition-induced distortion before its predictions
change. This is consistent with the finite-$D$ analysis in the main
text, where increasing $D$ reduces random-feature approximation error
but does not remove the structural interference induced by
superposition.

\begin{figure}[h]
    \centering
    \includegraphics[width=0.9\linewidth]
    {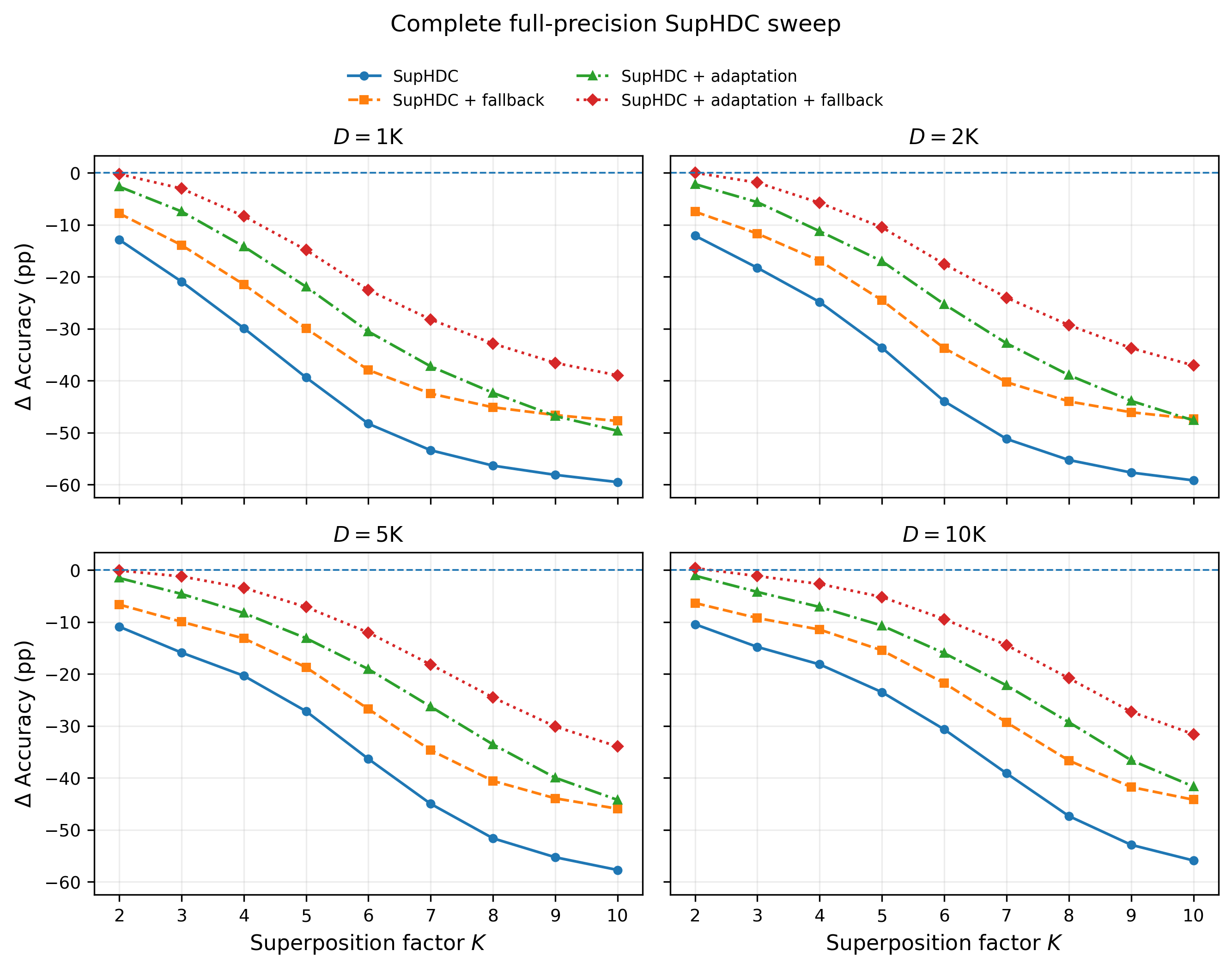}
    \caption{
    \textbf{Complete full-precision SupHDC sweep.}
    Each panel shows all four SupHDC variants over $K=2,\ldots,10$ for
    one representation dimension. The ordering among methods and the
    progressive degradation with increasing $K$ are consistent across
    $D=1$K, $2$K, $5$K, and $10$K, indicating that the trends observed
    in the main operating regime are not specific to a single
    representation dimension.
    }
    \label{fig:app_full_fp_grid}
\end{figure}

\textbf{Complete full-precision grid:}
For completeness, Figure~\ref{fig:app_full_fp_grid} reports all four
SupHDC variants for every evaluated representation dimension at full
precision.
The qualitative behavior of SupHDC is consistent across all evaluated representation dimensions, indicating that the method does not rely on operating specifically at $D=10{,}000$. Across $D=1K,2K,5K,$ and $10K$, moderate superposition remains usable, while increasing $D$ progressively improves tolerance to more aggressive $K$.
Across all four dimensions,
direct SupHDC exhibits the largest degradation, both adaptation and
fallback improve prediction result, and their combination provides
the strongest accuracy preservation. Likewise, increasing $K$
progressively increases the difficulty of the shared inference problem
at every $D$.

The complete grid also shows that the benefit of adaptation persists
when the representation itself is substantially smaller. At the same
time, larger $D$ shifts all of the curves toward lower accuracy
degradation, demonstrating that the two design choices address
different aspects of the problem: adaptation makes the slot-specific
readouts better matched to superposed inputs, whereas increasing $D$
improves the fidelity of the underlying random-feature representation.

\begin{figure}[h]
    \centering
    \includegraphics[width=0.8\linewidth]
    {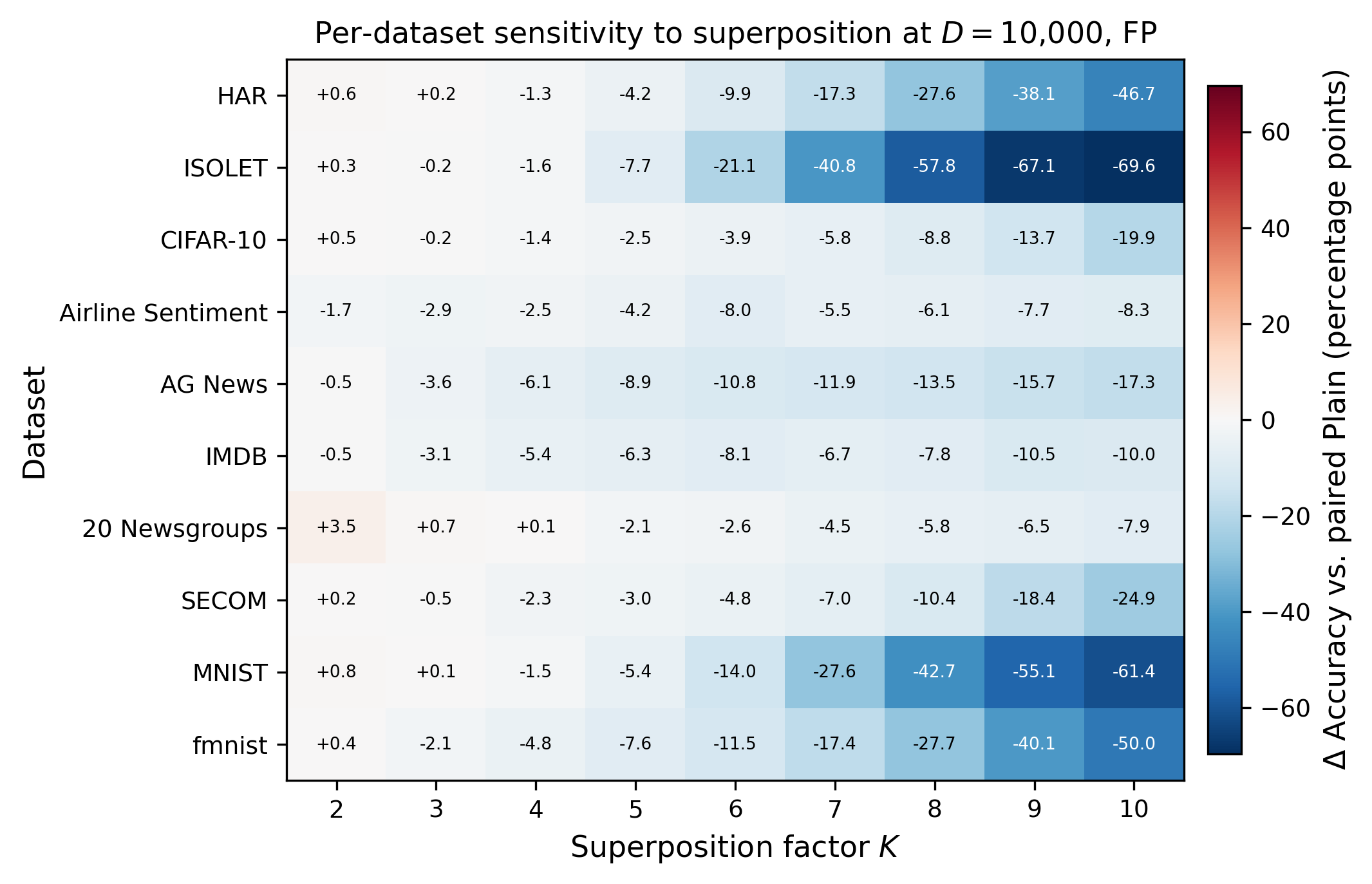}
    \caption{
    \textbf{Per-dataset sensitivity to the superposition factor.}
    Paired accuracy change for SupHDC with adaptation and $20\%$
    selective fallback at $D=10{,}000$ and full precision. Each cell
    reports the accuracy change in percentage points relative to the
    corresponding Plain HDC model, averaged over five paired seeds.
    Moderate superposition is broadly tolerated, whereas larger $K$
    produces increasingly heterogeneous degradation across workloads.
    }
    \label{fig:app_dataset_heatmap}
\end{figure}

\textbf{Dataset-level sensitivity to superposition:}
The aggregate curves summarize the average operating behavior, but the
difficulty of superposed inference is not uniform across datasets.
Figure~\ref{fig:app_dataset_heatmap} therefore reports the paired
accuracy change separately for each dataset at $D=10{,}000$ using
adaptation and $20\%$ fallback. In the moderate regime, the behavior is
broadly stable across tasks: at $K=2$, most datasets remain close to or
slightly above their paired Plain baseline, and degradation generally
remains limited through $K=3$--$4$. At larger $K$, however, dataset
sensitivity becomes much more pronounced.

In particular, ISOLET, MNIST, HAR, and Fashion-MNIST become increasingly
sensitive as $K$ enters the high-superposition regime, whereas datasets
such as Airline Sentiment, 20 Newsgroups, and IMDB degrade more
gradually. This heterogeneity explains the increasing cross-dataset
uncertainty visible in the aggregate curves at large $K$. It also
suggests that the appropriate superposition factor is application
dependent: $K$ acts as an inference-time efficiency knob, but the amount
of usable superposition depends on the classification margins and score
distortion tolerated by a particular workload.

Taken together, the extended results support three conclusions.
First, the favorable operating points reported in the main text are
part of a smooth and systematic accuracy--compute tradeoff rather than
isolated configurations. Second, adaptation and selective fallback are
complementary: adaptation improves the superposed readout without
additional inference encodings, while fallback selectively spends
additional computation on vulnerable queries. Third, representation
dimension and superposition factor provide distinct controls over the
system: larger $D$ improves tolerance to interference, while larger $K$
increases encoding amortization. In practice, moderate superposition
provides the most favorable balance; pushing $K$ substantially higher
continues to increase nominal compute savings, but with diminishing
returns and increasingly dataset-dependent accuracy degradation.

\section{Ablations of Adaptation and Selective Fallback}
\label{app:ablations}

We further examine the two mechanisms used to mitigate superposition
interference: superposition-aware prototype adaptation and margin-gated
selective fallback. These diagnostic ablations use representative
workloads spanning sensing, vision, and language, with $D=10{,}000$,
full-precision inference, and $K\in\{2,4\}$.

\begin{figure}[h]
    \centering
    \includegraphics[width=\linewidth]
    {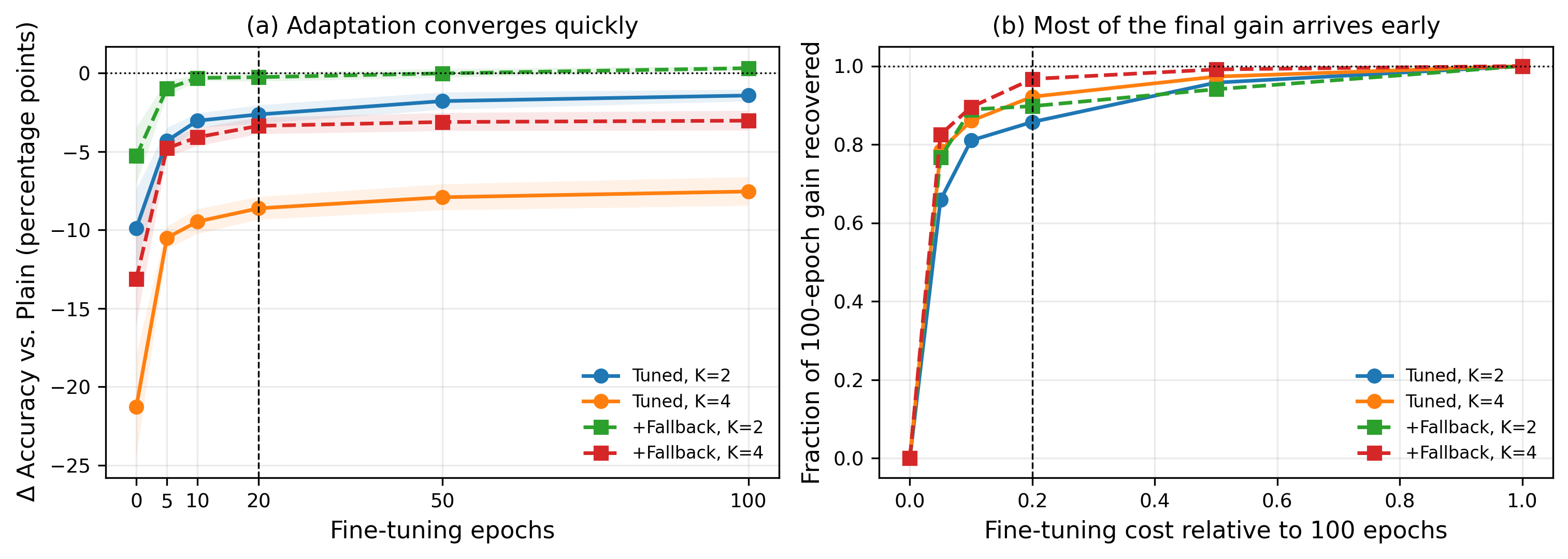}
    \caption{
    \textbf{Convergence of superposition-aware adaptation.}
    (a) Accuracy change relative to paired Plain HDC as the number of
    adaptation epochs increases. Most of the improvement occurs early,
    with diminishing gains after approximately 20 epochs.
    (b) Fraction of the final 100-epoch improvement recovered as a
    function of relative adaptation cost. Both adaptation alone and
    adaptation combined with fallback recover most of their final gain
    using only a fraction of the full adaptation budget.
    }
    \label{fig:app_adaptation_convergence}
\end{figure}

\textbf{Adaptation converges quickly:} Figure~\ref{fig:app_adaptation_convergence} studies the effect of the
number of superposition-aware adaptation epochs. Without adaptation,
superposed inference exhibits a substantial accuracy gap, particularly
at $K=4$. Most of this gap is recovered within the first few adaptation
epochs, after which the improvement gradually saturates. At 20 epochs,
the adapted models already recover approximately $86$--$92\%$ of the
improvement obtained after 100 epochs, depending on $K$. When combined
with selective fallback, approximately $90$--$97\%$ of the final gain is
already recovered at the same point.

These results indicate that the benefit of adaptation does not depend on
long optimization. The 100-epoch setting used in the main experiments
provides a conservative converged configuration, while substantially
shorter adaptation already captures most of the improvement.

\begin{figure}[h]
    \centering
    \includegraphics[width=\linewidth]
    {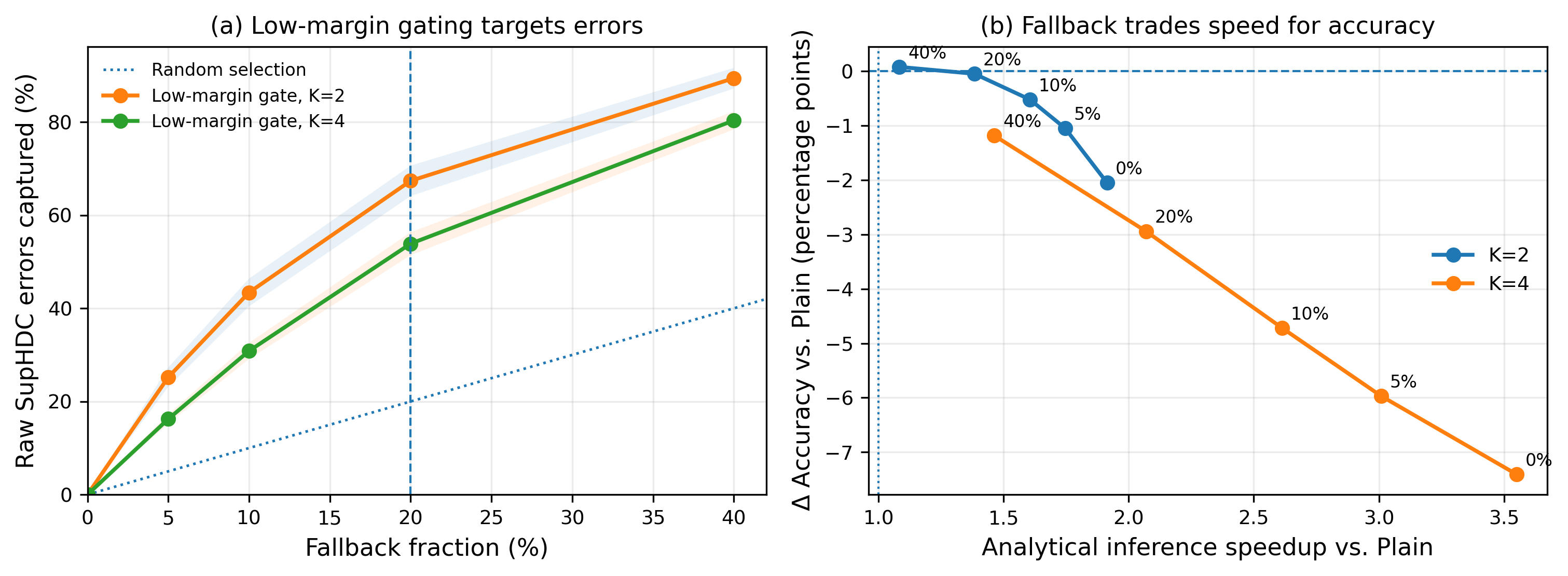}
    \caption{
    \textbf{Ablation of margin-gated selective fallback.}
    (a) Fraction of pre-fallback errors captured when queries are ranked
    by their top-two classification margin. Low-margin selection captures
    substantially more errors than random selection at the same fallback
    budget.
    (b) Accuracy--compute tradeoff as the fallback fraction is varied.
    Increasing fallback improves accuracy by selectively recomputing more
    uncertain queries, while giving back part of the analytical speedup.
    Percent labels indicate the fallback fraction.
    }
    \label{fig:app_fallback_ablation}
\end{figure}

\textbf{Low-margin fallback selectively targets errors:}
Figure~\ref{fig:app_fallback_ablation}(a) evaluates whether the
classification margin is a useful criterion for deciding which queries
to recompute. If fallback queries were selected uniformly at random,
recomputing a fraction $q$ of predictions would capture approximately
the same fraction $q$ of the pre-fallback errors. In contrast, selecting
the lowest-margin predictions captures errors at a substantially higher
rate. At a $20\%$ fallback fraction, the low-margin gate captures
approximately $67\%$ of the pre-fallback errors for $K=2$ and $54\%$
for $K=4$, compared with only $20\%$ expected under random selection.
Thus, the score margin provides a useful indicator of predictions that
are vulnerable to superposition interference.

Figure~\ref{fig:app_fallback_ablation}(b) shows the corresponding
accuracy--compute tradeoff as the fallback fraction is varied. Increasing
the fallback fraction progressively improves prediction accuracy because
more uncertain queries are recomputed through the clean inference path,
while reducing the overall analytical speedup. For $K=2$, relatively
small fallback fractions are sufficient to approach the Plain HDC
accuracy. For the more difficult $K=4$ setting, larger fallback fractions
recover substantially more accuracy while still retaining a meaningful
inference speedup. The $20\%$ fallback rate used in the main experiments
therefore represents a practical intermediate operating point rather
than a requirement of the method.

The two mechanisms address different aspects of superposition
interference. Adaptation improves the slot-specific classifiers so that
they better recognize class evidence directly from mixed
representations, and most of this benefit is obtained with relatively
few adaptation epochs. Selective fallback then uses the prediction
margin to identify the remaining difficult queries and spends additional
computation only where it is most useful. Together, they provide an
effective and tunable mechanism for improving prediction fidelity
without giving up the computational advantage of shared encoding.

\section{Detailed Computational and Storage Analysis}
\label{app:cost_storage}

We provide additional accounting for the computational and storage
tradeoffs of SupHDC. Let $d$ denote the input dimension, $D$ the
random-feature dimension, $C$ the number of classes, $K$ the
superposition factor, and $q$ the selective-fallback fraction.

\textbf{Analytical inference cost:}
For the complex RFF-HDC model used in our experiments, one
$d$-to-$D$ projection requires approximately $E=2Dd$ real FLOPs, while
scoring the resulting complex hypervector against $C$ class prototypes
requires approximately $R=4DC$ FLOPs. Processing $K$ queries
independently therefore costs
\begin{equation}
F_{\mathrm{Plain}} = K(E+R)
\label{eq:app_plain_cost}
\end{equation}

SupHDC replaces the $K$ independent projections with one shared
projection while retaining one prototype readout per slot:
\begin{equation}
F_{\mathrm{SupHDC}} = E + KR
\label{eq:app_sup_cost}
\end{equation}
Superposition-aware adaptation changes only the prototype banks and
does not add inference-time computation. With a fallback fraction $q$,
approximately $qK$ queries are additionally recomputed through the
clean inference path, giving
\begin{equation}
F_{\mathrm{SupHDC+FB}}
= E + KR + qK(E+R)
\label{eq:app_fb_cost}
\end{equation}

The signed-permutation keys require only lightweight permutation,
sign-flip, and summation operations in the original $d$-dimensional
space. As in the main paper, these operations are not included in the
analytical GFLOP count; their actual overhead is included in the
measured wall-clock evaluation.

Using the dataset-specific values of $d$ and $C$, this accounting gives
the analytical speedups reported in the main paper. The benefit approaches
$K\times$ when projection dominates the readout cost, while the
$KR$ term and selective recomputation lead to diminishing returns as
$K$ increases.

\textbf{Model storage cost:}
The computational saving comes with additional slot-specific classifier
state. The projection matrix $W$ is shared across all slots, whereas
SupHDC maintains one $C\times D$ prototype bank per slot. For adapted
SupHDC without fallback, deployment requires approximately
\begin{equation}
M_{\mathrm{SupHDC}}
\simeq M_W + K M_P + M_{\mathrm{key}}
\end{equation}
compared with $M_W+M_P$ for Plain HDC. When selective fallback is
enabled, both adapted and clean slot prototype banks are retained:
\begin{equation}
M_{\mathrm{SupHDC+FB}}
\simeq M_W + 2K M_P + M_{\mathrm{key}}
\end{equation}
Here $M_{\mathrm{key}}$ denotes the comparatively small storage for the
signed-permutation slot keys. We report deployment-minimal persistent
storage rather than temporary training or inference buffers.

\begin{figure}[h]
    \centering
    \includegraphics[width=0.72\linewidth]
    {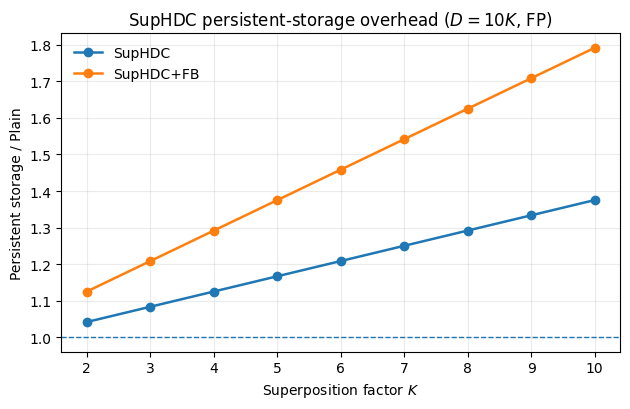}
    \caption{
    \textbf{Persistent storage overhead of SupHDC.}
    Mean persistent model storage relative to Plain HDC across the ten
    datasets at $D=10{,}000$ and full precision. The projection matrix
    is shared across slots, while slot-specific prototype banks account
    for the additional storage. Selective fallback retains an additional
    clean prototype bank for each slot.
    }
    \label{fig:app_storage}
\end{figure}

Figure~\ref{fig:app_storage} shows the mean storage ratio relative to
Plain HDC across the ten datasets at $D=10{,}000$ and full precision.
Storage grows gradually with $K$ because the dominant projection matrix
is shared. In the practical $K=2$--$4$ regime emphasized in the main
paper, adapted SupHDC requires only about $1.04$--$1.13\times$ the
persistent storage of Plain HDC. Retaining the additional clean
prototype banks for fallback raises this to approximately
$1.13$--$1.29\times$.

\begin{table}[h]
    \centering
    \caption{
    \textbf{Storage cost at representative operating points.}
    Values are averaged across the ten datasets at $D=10{,}000$ and
    full precision.
    }
    \label{tab:app_storage}
    \begin{tabular}{lccc}
        \toprule
        Method & $K$ & Storage (MiB) & Relative to Plain \\
        \midrule
        Plain       & -- & 25.19 & $1.00\times$ \\
        SupHDC      & 2  & 25.90 & $1.04\times$ \\
        SupHDC      & 4  & 27.30 & $1.13\times$ \\
        SupHDC + FB & 2  & 27.30 & $1.13\times$ \\
        SupHDC + FB & 4  & 30.11 & $1.29\times$ \\
        \bottomrule
    \end{tabular}
\end{table}

Overall, SupHDC trades a modest increase in persistent classifier
storage for a reduction in repeated high-dimensional encoding
computation. The projection matrix remains shared, so the storage
overhead is relatively small at the moderate superposition factors that
provide the strongest accuracy--compute tradeoffs. Selective fallback
requires additional computation and a clean prototype bank, but provides
a tunable mechanism for recovering prediction fidelity.

\section{Additional Edge Deployment Results}
\label{app:edge_results}

The main paper reports representative Raspberry Pi 5 operating points
for adapted SupHDC with and without $20\%$ selective fallback. Here, we
further examine whether the analytical cost model continues to reflect
measured hardware behavior as the fallback fraction is varied. We
evaluate $q\in\{0,0.05,0.10,0.20\}$ for $K\in\{2,4\}$ at
$D=10{,}000$ and full precision. As in the main deployment experiment,
speedups are aggregated across HAR, ISOLET, CIFAR-10, and AG News using
the geometric mean.

\begin{figure}[h]
    \centering
    \includegraphics[width=0.7\linewidth]
    {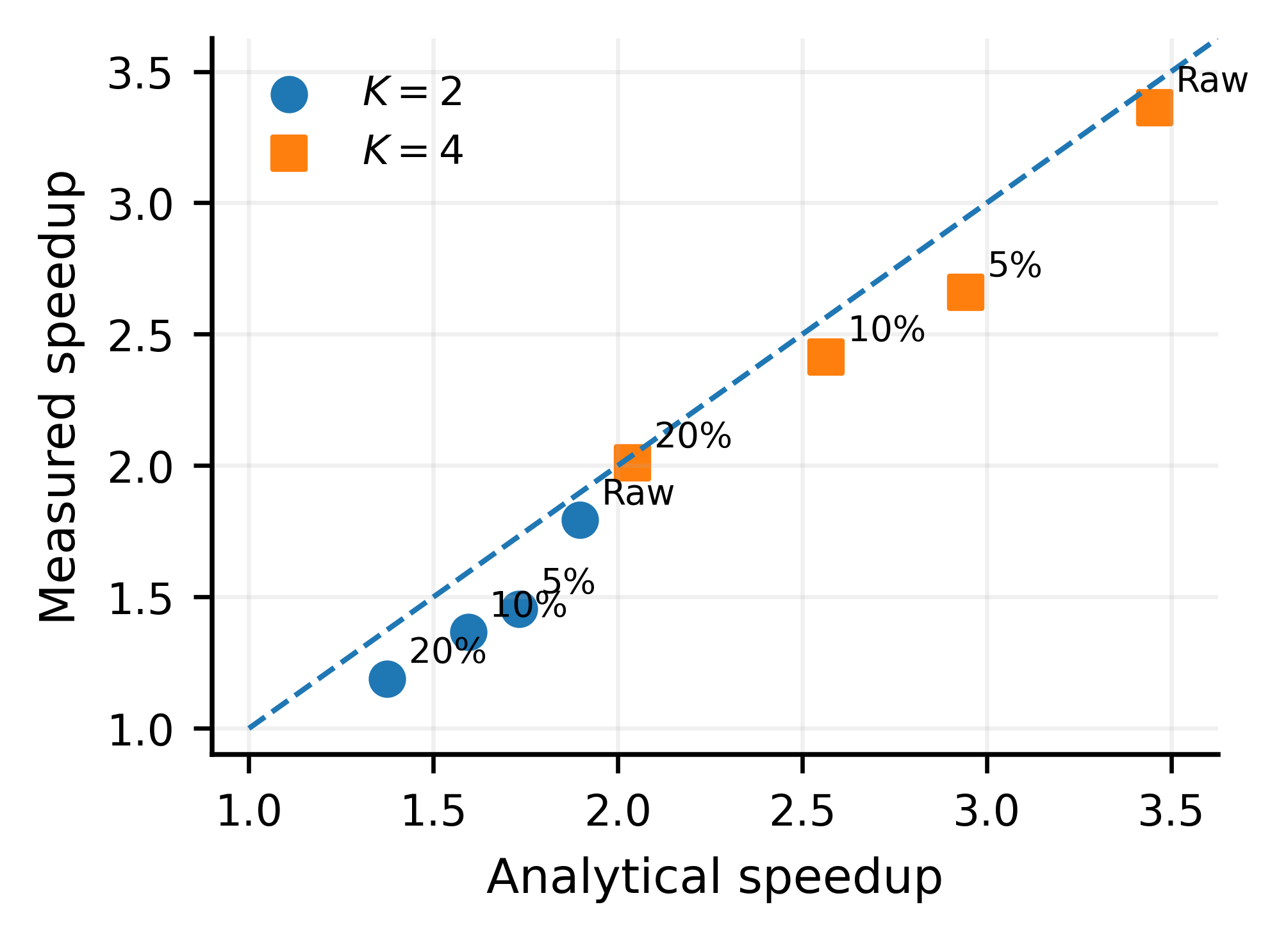}
    \caption{
    \textbf{Analytical versus measured Raspberry Pi 5 speedup.}
    Each point corresponds to adapted SupHDC with fallback fraction
    $q\in\{0,5,10,20\}\%$ for $K=2$ or $K=4$.
    ``Raw'' denotes the adapted SupHDC model without selective fallback
    ($q=0$), rather than unadapted direct superposition.
    Speedups are aggregated across HAR, ISOLET, CIFAR-10, and AG News.
    The dashed line denotes ideal agreement between analytical and measured
    speedup. Increasing fallback reduces acceleration because more
    low-confidence queries are individually recomputed.
    }
    \label{fig:app_pi_analytical_measured}
\end{figure}

Figure~\ref{fig:app_pi_analytical_measured} compares measured Raspberry
Pi speedup against the corresponding analytical prediction. In the
figure, ``Raw'' refers to adapted SupHDC before selective fallback
($q=0$). Across both superposition factors and all evaluated fallback
fractions, measured speedup follows the analytical trend closely.
Without fallback, $K=2$
achieves $1.79\times$ measured speedup compared with $1.90\times$
analytically, while $K=4$ achieves $3.36\times$ compared with
$3.45\times$. As the fallback fraction increases, both analytical and
measured speedup move progressively toward $1\times$ because a larger
fraction of queries is individually re-encoded through the clean path.

At the default $20\%$ fallback setting, measured speedups are
$1.19\times$ for $K=2$ and $2.01\times$ for $K=4$, compared with
analytical predictions of $1.38\times$ and $2.04\times$, respectively.
The remaining gap reflects implementation and system overheads that are
not captured by the hardware-independent FLOP model. Overall, the
agreement across the complete fallback sweep supports the use of the
analytical cost model for characterizing SupHDC's inference tradeoffs.

\section{Discussion and Limitations}
\label{app:discussion}

SupHDC suggests that high-dimensional redundancy can serve not only as
representation robustness, but also as shared inference capacity.
Rather than recovering each individual hypervector exactly, multiple
queries can share one encoding computation as long as the class evidence
needed for prediction is sufficiently preserved.

This makes SupHDC complementary to existing efficiency techniques.
Reducing $D$, lowering numerical precision, or specializing hardware makes
each encoding cheaper; SupHDC instead reduces how many independent
encodings are required. The Raspberry Pi results further show that this
algorithmic saving can translate into real wall-clock acceleration.
More broadly, this suggests superposed inference as an additional design
dimension for HDC and related kernel models.

Several limitations also point to natural directions for future work:

\begin{itemize}
    \item \textbf{The useful superposition factor $K$ is workload dependent:}
    Different datasets exhibit different tolerance to superposition, and the
    current method selects $K$ empirically. A useful direction is to develop
    diagnostics that predict how well a workload will respond to superposed
    inference, or directly select an appropriate $K$ from properties such
    as classification margin or score distortion.

    \item \textbf{Random grouping may not be optimal:}
    Our experiments form superposed groups from randomly paired queries.
    However, interference depends on the interaction among samples within a
    group, suggesting that interference-aware grouping or scheduling could
    further improve the accuracy--compute tradeoff.

    \item \textbf{The current study focuses on projection-based HDC:}
    Projection-based encoders are prominent in HDC and provide a natural
    connection to kernel methods, but they do not cover the
    full range of HDC/VSA architectures. Extending superposed inference to
    other encoders and representation schemes remains an important direction.
\end{itemize}

\end{document}

%% file: math_commands.tex
\usepackage{amsmath,amsfonts,bm}

\def\eqref#1{equation~\ref{#1}}
\def\1{\bm{1}}

\DeclareMathAlphabet{\mathsfit}{\encodingdefault}{\sfdefault}{m}{sl}
\SetMathAlphabet{\mathsfit}{bold}{\encodingdefault}{\sfdefault}{bx}{n}